# Unifying Class-Based Representation Formalisms

**Diego Calvanese**                                    CALVANESE@DIS.UNIROMA1.IT
**Maurizio Lenzerini**                                 LENZERINI@DIS.UNIROMA1.IT
**Daniele Nardi**                                      NARDI@DIS.UNIROMA1.IT
*Dipartimento di Informatica e Sistemistica*
*Università di Roma "La Sapienza"*
*Via Salaria 113, I-00198 Roma, Italy*

## Abstract

The notion of class is ubiquitous in computer science and is central in many formalisms for the representation of structured knowledge used both in knowledge representation and in databases. In this paper we study the basic issues underlying such representation formalisms and single out both their common characteristics and their distinguishing features. Such investigation leads us to propose a unifying framework in which we are able to capture the fundamental aspects of several representation languages used in different contexts. The proposed formalism is expressed in the style of description logics, which have been introduced in knowledge representation as a means to provide a semantically well-founded basis for the structural aspects of knowledge representation systems. The description logic considered in this paper is a subset of first order logic with nice computational characteristics. It is quite expressive and features a novel combination of constructs that has not been studied before. The distinguishing constructs are number restrictions, which generalize existence and functional dependencies, inverse roles, which allow one to refer to the inverse of a relationship, and possibly cyclic assertions, which are necessary for capturing real world domains. We are able to show that it is precisely such combination of constructs that makes our logic powerful enough to model the essential set of features for defining class structures that are common to frame systems, object-oriented database languages, and semantic data models. As a consequence of the established correspondences, several significant extensions of each of the above formalisms become available. The high expressiveness of the logic we propose and the need for capturing the reasoning in different contexts forces us to distinguish between unrestricted and finite model reasoning. A notable feature of our proposal is that reasoning in both cases is decidable. We argue that, by virtue of the high expressive power and of the associated reasoning capabilities on both unrestricted and finite models, our logic provides a common core for class-based representation formalisms.

## 1. Introduction

In many fields of computer science we find formalisms for the representation of objects and classes (Motschnig-Pitrik & Mylopoulous, 1992). Generally speaking, an object denotes an element of the domain of interest, and a *class* denotes a set of objects with common characteristics. We use the term "class-based representation formalism" to refer to a formalism that allows one to express several kinds of relationships and constraints (e.g., subclass constraints) holding among the classes that are meaningful in a set of applications. Moreover, class-based formalisms aim at taking advantage of the class structure in order to provide various information, such as whether a class is consistent, i.e., it admits at least one object, whether a class is a subclass of another class, and more generally, whether a given constraint holds between a given set of classes. From the above characterization, it should be clear that the formalisms referred to in this paper deal only with the structural aspects of objects and classes, and do not include any features for the specification of behavioral properties of objects.

Three main families of class-based formalisms are identified in this paper. The first one comes from knowledge representation and in particular from the work on semantic networks and frames (see for example Lehmann, 1992; Sowa, 1991). The second one originates in the field of databases and in particular from the work on semantic data models (see for example Hull & King, 1987). The third one arises from the work on types in programming languages and object-oriented systems (see for example Kim & Lochovsky, 1989).

In the past there have been several attempts to establish relationships among the various families of class-based formalisms (see Section 6 for a brief survey). The proposed solutions are not fully general and a formalism capturing both the modeling constructs and the reasoning techniques for all the above families is still missing. In this paper we address this problem by proposing a class-based representation formalism, based on description logics (Brachman & Levesque, 1984; Schmidt-Schauß & Smolka, 1991; Donini, Lenzerini, Nardi, & Schaerf, 1996), and by using it for comparing other formalisms.

In description logics, structured knowledge is described by means of so called *concepts* and *roles*, which denote unary and binary predicates, respectively. Starting from a set of atomic symbols one can build complex concept and role expressions by applying suitable constructors which characterize a description logic. Formally, concepts are interpreted as subsets of a domain and roles as binary relations over that domain, and all constructs are equipped with a precise set-theoretic semantics. The most common constructs include boolean operations on concepts, and quantification over roles. For example, the concept `Person ⊓ ∀child.Male`, denotes the set of individuals that are instances of the concept `Person` and are connected through the role `child` only to instances of the concept `Male`, while the concept `∃child` denotes all individuals that are connected through the role `child` to some individual. Further constructs that have been considered important include more general forms of quantification, number restrictions, which allow one to state limits on the number of connections that an individual may have via a certain role, and constructs on roles, such as intersection, concatenation and inverse. A description logic knowledge base, expressing the intensional knowledge about the modeled domain, is built by stating inclusion assertions between concepts, which have to be satisfied by the models of the knowledge base. The assertions are used to specify necessary and/or necessary and sufficient conditions for individuals to be instances of certain concepts. Reasoning on such knowledge bases includes the detection of inconsistencies in the knowledge base itself, determining whether a concept can be populated in a model of the knowledge base, and checking subsumption, i.e., whether all instances of a concept are necessarily also instances of another concept in all models of the knowledge base.

In this paper we propose a description logic called ALUNI, which is quite expressive and includes a novel combination of constructs, including number restrictions, inverse roles, and inclusion assertions with no restrictions on cycles. Such features make ALUNI powerful enough to provide a unified framework for frame systems, object-oriented languages, and semantic data models. We show this by establishing a precise correspondence with a frame-based language in the style of the one proposed by Fikes and Kehler (1985), with the Entity-Relationship model (Chen, 1976), and with an object-oriented language in the style of the one introduced by Abiteboul and Kanellakis (1989). More specifically, we identify the most relevant features to model classes in each of the cited settings and show that a specification in any of those class-based formalisms can be equivalently expressed as a knowledge base in ALUNI. In this way, we are able to identify which are the commonalities among the families and which are the specificities of each family. Therefore, even though there are specific features of every family that are not addressed by ALUNI, we are able to show that the formalism proposed in this paper provides important features that are currently missing in each family, although their relevance has often been stressed. In this sense, our unifying framework points out possible developments for the languages belonging to all the three families.

One fundamental reason for regarding ALUNI as a unifying framework for class-based representation formalisms is that reasoning in ALUNI is hard, but nonetheless decidable, as shown by Calvanese, Lenzerini, and Nardi (1994), Calvanese (1996c). Consequently, the language features arising from different frameworks to build class-based representations are not just given a common semantic account, but are combined in a more expressive setting where one retains the capability of solving reasoning tasks. The combination of constructs included in the language makes it necessary to distinguish between reasoning with respect to finite models, i.e., models with a finite domain, and reasoning with respect to unrestricted models. Calvanese (1996c) devises suitable techniques for both unrestricted and finite model reasoning, that enable for reasoning in the different contexts arising from assuming a finite domain, as it is often the case in the field of databases, or assuming that a domain can also be infinite. In the paper, we discuss the results on reasoning in ALUNI, and compare them with other results on reasoning in class-based representation formalisms.

Summarizing, our framework provides an adequate expressive power to account for the most significant features of the major families of class-based formalisms. Moreover, it is equipped with suitable techniques for reasoning in both finite and unrestricted models. Therefore, we believe that ALUNI captures the essential core of the class-based representation formalisms belonging to all three families mentioned above.

The paper is organized as follows. In the next section we present our formalism and in Sections 3, 4, and 5 we discuss three families of class-based formalisms, namely, frame languages, semantic data models, and object-oriented data models, showing that their basic features are captured by knowledge bases in ALUNI. The final sections contain a review of related work, including a discussion of reasoning in ALUNI and class-based formalism, and some concluding remarks.

## 2. A Unifying Class-Based Representation Language

In this section, we present ALUNI, a class-based formalism in the style of *description logics* (DLs) (Brachman & Levesque, 1984; Schmidt-Schauß & Smolka, 1991; Donini et al., 1996; Donini, Lenzerini, Nardi, & Nutt, 1997). In DLs the domain of interest is modeled by means of *concepts* and *roles*, which denote classes and binary relations, respectively. Generally speaking, a DL is formed by three basic components:

- A *description language*, which specifies how to construct complex concept and role expressions (also called simply concepts and roles), by starting from a set of atomic symbols and by applying suitable constructors. It is the set of allowed constructs that characterizes the description language.

| Construct | Syntax | Semantics |
|---|---|---|
| atomic concept | $A$ | $A^{\mathcal{I}} \subseteq \Delta^{\mathcal{I}}$ |
| atomic negation | $\neg A$ | $\Delta^{\mathcal{I}} \setminus A^{\mathcal{I}}$ |
| conjunction | $C_1 \sqcap C_2$ | $C_1^{\mathcal{I}} \cap C_2^{\mathcal{I}}$ |
| disjunction | $C_1 \sqcup C_2$ | $C_1^{\mathcal{I}} \cup C_2^{\mathcal{I}}$ |
| universal quantification | $\forall R.C$ | $\{o \mid \forall o' . \, (o, o') \in R^{\mathcal{I}} \to o' \in C^{\mathcal{I}}\}$ |
| number restrictions | $\exists^{\geq n} R$ | $\{o \mid \sharp\{o' \mid (o, o') \in R^{\mathcal{I}}\} \geq n\}$[1] |
|  | $\exists^{\leq n} R$ | $\{o \mid \sharp\{o' \mid (o, o') \in R^{\mathcal{I}}\} \leq n\}$ |
| atomic role | $P$ | $P^{\mathcal{I}} \subseteq \Delta^{\mathcal{I}} \times \Delta^{\mathcal{I}}$ |
| inverse role | $P^-$ | $\{(o, o') \mid (o', o) \in P^{\mathcal{I}}\}$ |

Table 1: Syntax and semantics of $\mathcal{ALUNI}$

- A *knowledge specification mechanism*, which specifies how to construct a DL knowledge base, in which properties of concepts and roles are asserted.

- A set of *basic reasoning tasks* provided by the DL.

In the rest of the section we describe the specific form that these three components assume in ALUNI.

## 2.1 The Description Language of ALUNI

In the description language of ALUNI, called $\mathcal{ALUNI}$, *concepts* and *roles* are formed according to the syntax shown in Table 1, where $A$ denotes an atomic concept, $P$ an atomic role, $C$ an arbitrary concept expression, $R$ an arbitrary role expression, and $n$ a nonnegative integer. To increase readability of concept expressions, we also introduce the following abbreviations:

$$\top \equiv A \sqcup \neg A, \quad \text{for some atomic concept } A$$
$$\bot \equiv A \sqcap \neg A, \quad \text{for some atomic concept } A$$
$$\exists R \equiv \exists^{\geq 1} R$$
$$\exists^{=n} R \equiv \exists^{\leq n} R \sqcap \exists^{\geq n} R$$

Concepts are interpreted as subsets of a domain and roles as binary relations over that domain. Intuitively, $\neg A$ represents the *negation* of an atomic concept, and is interpreted as the complement with respect to the domain of interpretation. $C_1 \sqcap C_2$ represents the *conjunction* of two concepts and is interpreted as set intersection, while $C_1 \sqcup C_2$ represents *disjunction* and is interpreted as set union. Consequently, $\top$ represents the whole domain, and $\perp$ the empty set. $\forall R.C$ is called *universal quantification over roles* and is used to denote those elements of the interpretation domain that are connected through role $R$ only to instances of the concept $C$. $\exists^{\geq n} R$ and $\exists^{\leq n} R$ are called *number restrictions*, and impose on their instances restrictions on the minimum and maximum number of objects they are connected to through role $R$. $P^-$, called the *inverse* of role $P$, represents the inverse of the binary relation denoted by $P$.

---

1. $\sharp S$ denotes the cardinality of a set $S$.

More formally, an interpretation $\mathcal{I} = (\Delta^{\mathcal{I}}, \cdot^{\mathcal{I}})$ consists of an *interpretation domain* $\Delta^{\mathcal{I}}$ and an *interpretation function* $\cdot^{\mathcal{I}}$ that maps every concept $C$ to a subset $C^{\mathcal{I}}$ of $\Delta^{\mathcal{I}}$ and every role $R$ to a subset $R^{\mathcal{I}}$ of $\Delta^{\mathcal{I}} \times \Delta^{\mathcal{I}}$ according to the semantic rules specified in Table 1. The sets $C^{\mathcal{I}}$ and $R^{\mathcal{I}}$ are called the *extensions* of $C$ and $R$ respectively.

**Example 2.1** Consider the concept expression

$$\forall \texttt{enrolls.Student} \sqcap \exists^{\geq 2}\texttt{enrolls} \sqcap \exists^{\leq 30}\texttt{enrolls} \sqcap$$
$$\forall \texttt{teaches}^-.(\texttt{Professor} \sqcup \texttt{GradStudent}) \sqcap \exists^{=1}\texttt{teaches}^- \sqcap$$
$$\neg\texttt{AdvCourse}$$

specifying the constraints for an object to be a university course. The expression reflects the fact that each course enrolls only students, and restrictions on the minimum and maximum number of enrolled students. By using the role `teaches` and the inverse constructor we can state the property that each course is taught by exactly one individual, who is either a professor or a graduate student. Finally, negation is used to express disjointness from the concept denoting advanced courses. ∎

## 2.2 Knowledge Bases in ALUNI

An ALUNI *knowledge base*, which expresses the knowledge about classes and relations of the modeled domain, is formally defined as a triple $\mathcal{K} = (\mathcal{A}, \mathcal{P}, \mathcal{T})$, where $\mathcal{A}$ is a finite set of atomic concepts, $\mathcal{P}$ is a finite set of atomic roles, and $\mathcal{T}$ is a finite set of so called *inclusion assertions*. Each such assertion has the form

$$A \;\dot{\preceq}\; C$$

where $A$ is an atomic concept and $C$ an arbitrary concept expression. Such an inclusion assertion states by means of the concept $C$ necessary properties for an element of the domain in order to be an instance of the atomic concept $A$. Formally, an interpretation $\mathcal{I}$ *satisfies* the inclusion assertion $A \dot{\preceq} C$ if $A^{\mathcal{I}} \subseteq C^{\mathcal{I}}$. An interpretation $\mathcal{I}$ is a *model* of a knowledge base $\mathcal{K}$ if it satisfies all inclusion assertions in $\mathcal{K}$. A *finite model* is a model with finite domain.

**Example 2.1 (cont.)** The assertion

$$\texttt{Course} \;\dot{\preceq}\; \forall\texttt{enrolls.Student} \sqcap \exists^{\geq 2}\texttt{enrolls} \sqcap \exists^{\leq 30}\texttt{enrolls} \sqcap$$
$$\forall\texttt{teaches}^-.(\texttt{Professor} \sqcup \texttt{GradStudent}) \sqcap \exists^{=1}\texttt{teaches}^-$$

makes use of a complex concept expression to state necessary conditions for an object to be an instance of the concept `Course`. ∎

In ALUNI no restrictions are imposed on the form that the inclusion assertions may assume. In particular we do not rule out *cyclic assertions*, i.e., assertions in which the concept expression on the right hand side refers, either directly or indirectly via other assertions, to the atomic concept on the left hand side. In the presence of cyclic assertions different semantics may be adopted (Nebel, 1991). The one defined above, called *descriptive semantics*, accepts all interpretations that satisfy the assertions in the knowledge base, and hence interprets assertions as constraints on the domain to be modeled. For inclusion assertions, descriptive semantics has been claimed to provide the most intuitive results (Buchheit, Donini, Nutt, & Schaerf, 1998). Alternative semantics which have been proposed are based on fixpoint constructions (Nebel, 1991; Schild, 1994; De Giacomo & Lenzerini, 1994b), and hence allow to define in a unique way the interpretation of concepts.

In general, cycles in the knowledge base increase the complexity of reasoning (Nebel, 1991; Baader, 1996; Calvanese, 1996b) and require a special treatment by reasoning procedures (Baader, 1991; Buchheit, Donini, & Schaerf, 1993). For this reason, many DL based systems assume the knowledge base to be acyclic (Brachman, McGuinness, Patel-Schneider, Alperin Resnick, & Borgida, 1991; Bresciani, Franconi, & Tessaris, 1995). However, this assumption is unrealistic in practice, and cycles are definitely necessary for a correct modeling in many application domains. Indeed, the use of cycles is allowed in all data models used in databases, and, as shown in the following sections, in order to capture their semantics in ALUNI the possibility of using cyclic assertions is fundamental.

Besides inclusion assertions, some DL based systems also make use of equivalence assertions (Buchheit et al., 1993), which express both necessary and sufficient conditions for an object to be an instance of a concept. Although this possibility is ruled out in ALUNI, this does not limit its ability of capturing both frame based systems and database models, where the constraints that can be expressed correspond naturally to inclusion assertions.

## 2.3 Reasoning in ALUNI

The basic tasks we consider when reasoning over an ALUNI knowledge base are concept consistency and concept subsumption:

- *Concept consistency* is the problem of deciding whether a concept $C$ is *consistent* in a knowledge base $\mathcal{K}$ (written as $\mathcal{K} \not\models C \preceq \bot$), i.e., whether $\mathcal{K}$ admits a model $\mathcal{I}$ such that $C^{\mathcal{I}} \neq \emptyset$.

- *Concept subsumption* is the problem of deciding whether a concept $C_1$ *is subsumed by* a concept $C_2$ in a knowledge base $\mathcal{K}$ (written as $\mathcal{K} \models C_1 \preceq C_2$), i.e., whether $C_1^{\mathcal{I}} \subseteq C_2^{\mathcal{I}}$ for each model $\mathcal{I}$ of $\mathcal{K}$.

The inclusion of number restrictions and inverse roles in $\mathcal{ALUNI}$ and the ability in ALUNI of using arbitrary, possibly cyclic inclusion assertions allows one to construct a knowledge base in which a certain concept is consistent but has necessarily an empty extension in all finite models of the knowledge base. Similarly, a subsumption relation between two concepts may hold only if infinite models of the knowledge base are ruled out and only finite models are considered.

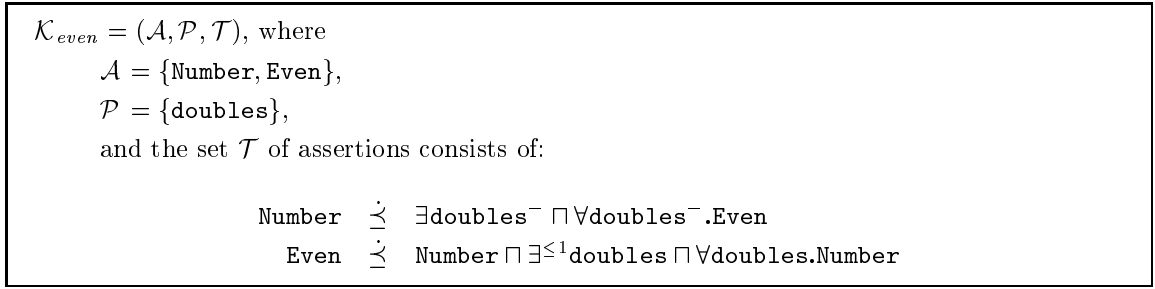

$\mathcal{K}_{even} = (\mathcal{A}, \mathcal{P}, \mathcal{T})$, where

    $\mathcal{A} = \{\text{Number}, \text{Even}\}$,

    $\mathcal{P} = \{\text{doubles}\}$,

    and the set $\mathcal{T}$ of assertions consists of:

        Number $\;\dot{\preceq}\;$ $\exists\text{doubles}^- \sqcap \forall\text{doubles}^-.\text{Even}$

        Even $\;\dot{\preceq}\;$ $\text{Number} \sqcap \exists^{\leq 1}\text{doubles} \sqcap \forall\text{doubles}.\text{Number}$

Figure 1: An ALUNI knowledge base with two concepts that are equivalent in all finite models

**Example 2.2** Let $\mathcal{K}_{even}$ be the knowledge base shown in Figure 1. Intuitively, the assertions in $\mathcal{K}_{even}$ state that for each number there is an even number which doubles it, and that all numbers which double it are even. Each even number is a number, doubles at most one number, and doubles only numbers. Observe that for any model $\mathcal{I}$ of $\mathcal{K}_{even}$, the universal quantifications together with the functionality of doubles in the assertions imply that $\sharp\text{Even}^{\mathcal{I}} \geq \sharp\text{Number}^{\mathcal{I}}$, while the direct inclusion assertion between Even and Number implies that $\sharp\text{Even}^{\mathcal{I}} \leq \sharp\text{Number}^{\mathcal{I}}$. Therefore, the two concepts have the same cardinality, and since one is a sub-concept of the other, if the domain is finite, their extensions coincide. This does not necessarily hold for infinite domains. In fact, the names we have chosen suggest already an infinite model of the knowledge base in which Number and Even are interpreted differently. The model is obtained by taking the natural numbers as domain, and interpreting Number as the whole domain, Even as the even numbers, and doubles as the set $\{(2n, n) \mid n \geq 0\}$. ∎

The example above shows that ALUNI does not have the *finite model property*, which states that if a concept is consistent in a knowledge base then the knowledge base admits a *finite* model in which the concept has a nonempty extension. Therefore, it is important to distinguish between reasoning with respect to unrestricted models and reasoning with respect to finite models. We call *(unrestricted) concept consistency* (written as $\mathcal{K} \not\models_u C \preceq \bot$) and *(unrestricted) concept subsumption* (written as $\mathcal{K} \models_u A \preceq C$) the reasoning tasks as described above, i.e., carried out without restricting the attention to finite models. The corresponding reasoning tasks carried out by considering finite models only, are called *finite concept consistency* (written as $\mathcal{K} \not\models_f C \preceq \bot$) and *finite concept subsumption* (written as $\mathcal{K} \models_f A \preceq C$).

**Example 2.2 (cont.)** Summing up the previous considerations, we can say that Number is not subsumed by Even in $\mathcal{K}_{even}$, i.e., $\mathcal{K}_{even} \not\models_u \text{Number} \preceq \text{Even}$, but is finitely subsumed, i.e., $\mathcal{K}_{even} \models_f \text{Number} \preceq \text{Even}$. Equivalently $\text{Number} \sqcap \neg\text{Even}$ is consistent in $\mathcal{K}_{even}$, i.e., $\mathcal{K}_{even} \not\models_u \text{Number} \sqcap \neg\text{Even} \preceq \bot$, but is not finitely consistent, i.e., $\mathcal{K}_{even} \models_f \text{Number} \sqcap \neg\text{Even} \preceq \bot$. ∎

A distinguishing feature of ALUNI is that reasoning both in the finite and in the unrestricted case is decidable. In particular, unrestricted concept satisfiability and concept subsumption are decidable in deterministic exponential time (De Giacomo & Lenzerini, 1994a; Calvanese et al., 1994), and since reasoning in strict sublanguages of ALUNI is already EXPTIME-hard (Calvanese, 1996c), the known algorithms are computationally optimal. Finite concept consistency in ALUNI is also decidable in deterministic exponential time while finite concept subsumption (in the general case) is decidable in deterministic double exponential time (Calvanese, 1996c). A more precise discussion on the methods for reasoning in ALUNI is provided in Section 6.2, while a detailed account of the adopted algorithms and an analysis of their computational complexity is presented by Calvanese (1996c).

In the next sections we show how the two forms of reasoning with respect to unrestricted and finite models, capture the reasoning tasks that are typically considered in different formalisms for the structured representation of knowledge.

## 3. Frame Based Systems

Frame languages are based on the idea of expressing knowledge by means of *frames*, which are structures representing classes of objects in terms of the properties that their instances must satisfy. Such properties are defined by the frame *slots*, that constitute the items of a frame definition. Since the 70s a large number of frame systems have been developed, with different goals and different features. DLs bear a close relationship with the KL-ONE family of frame systems (Woods & Schmolze, 1992). However, here we would like to consider frame systems from a more general perspective, as discussed for example by Karp (1992), Karp, Myers, and Gruber (1995), and establish the correspondence with ALUNI knowledge bases in this context.

We remark that we are restricting our attention to those aspects that are related to the taxonomic structure. Moreover, as discussed below, we consider assertional knowledge bases, where intensional knowledge is characterized in terms of inclusion assertions rather than definitions. In addition, we do not consider those features that cannot be captured in a first-order framework, such as default values in the slots, attached procedures, and the specification of overriding inheritance policies. Some of the issues concerning the modeling of these aspects in DLs are addressed by Donini, Lenzerini, Nardi, Nutt, and Schaerf (1994), Donini, Nardi, and Rosati (1995), within a modal nonmonotonic extension of DLs.

### 3.1 Syntax of Frame Based Systems

To make the correspondence precise, we need to fix syntax and semantics for the frame systems we consider. Unfortunately, there is no accepted standard and we have chosen to use here basically the notation adopted by Fikes and Kehler (1985), which is used also in the KEE[2] system.

**Definition 3.1** A *frame knowledge base*, denoted by $\mathcal{F}$, is formed by a set of *frame* and *slot names*, and is constituted by a set of *frame definitions* of the following form:

$$\textbf{Frame}: \ F \ \textbf{in KB} \ \mathcal{F} \quad E,$$

---

2. KEE is a trademark of Intellicorp. Note that a KEE user does not directly specify her knowledge base in this notation, but is allowed to define frames interactively via the graphical system interface.

```
Frame: Course in KB University            Frame: BasCourse in KB University
   MemberSlot: enrolls                       SuperClasses: Course
      ValueClass: Student                    MemberSlot: taughtby
      Cardinality.Min: 2                         ValueClass: Professor
      Cardinality.Max: 30
   MemberSlot: taughtby                    Frame: Professor in KB University
      ValueClass: (UNION GradStudent
                        Professor)         Frame: Student in KB University
      Cardinality.Min: 1
      Cardinality.Max: 1                   Frame: GradStudent in KB University
                                              SuperClasses: Student
Frame: AdvCourse in KB University             MemberSlot: degree
   SuperClasses: Course                        ValueClass: String
   MemberSlot: enrolls                         Cardinality.Min: 1
      ValueClass: (INTERSECTION                Cardinality.Max: 1
                  GradStudent
                  (NOT Undergrad))        Frame: Undergrad in KB University
      Cardinality.Max: 20                     SuperClasses: Student
```

Figure 2: A KEE knowledge base

where $E$ is a *frame expression*, i.e., an expression formed according to the following syntax:

$$
\begin{aligned}
E \longrightarrow \ &\textbf{SuperClasses}: \ F_1, \ldots, F_h \\
&\textbf{MemberSlot}: \ S_1 \\
&\quad \textbf{ValueClass}: \ H_1 \\
&\quad \textbf{Cardinality.Min}: \ m_1 \\
&\quad \textbf{Cardinality.Max}: \ n_1 \\
&\ldots \\
&\textbf{MemberSlot}: \ S_k \\
&\quad \textbf{ValueClass}: \ H_k \\
&\quad \textbf{Cardinality.Min}: \ m_k \\
&\quad \textbf{Cardinality.Max}: \ n_k
\end{aligned}
$$

$F$ and $S$ denote frame and slot names, respectively, $m$ and $n$ denote positive integers, and $H$ denotes *slot constraint*, which can be specified as follows:

$$
\begin{aligned}
H \longrightarrow \ &F \ | \\
&(\text{INTERSECTION } H_1 \ H_2) \ | \\
&(\text{UNION } H_1 \ H_2) \ | \\
&(\text{NOT } H)
\end{aligned}
$$

∎

For readers that are familiar with the KEE system, we point out that we omit the specification of the sub-classes for a frame present in KEE, since it can be directly derived from the specification of the super-classes.

**Example 3.2** Figure 2 shows a simple example of a knowledge base modeling the situation at an university expressed in the frame language we have presented. The frame `Course` represents courses which enroll students and are taught either by graduate students or professors. Cardinality restrictions are used to impose a minimum and maximum number of students that may be enrolled in a course, and to express that each course is taught by exactly one individual. The frame `AdvCourse` represents courses which enroll only graduate students, i.e., students who already have a degree. Basic courses, on the other hand, may be taught only by professors. ∎

## 3.2 Semantics of Frame Based Systems

To give semantics to a set of frame definitions we resort to their interpretation in terms of first-order predicate calculus (Hayes, 1979). According to such interpretation, frame names are treated as unary predicates, while slots are considered binary predicates.

A frame expression $E$ is interpreted as a predicate logic formula $E(x)$, which has one free variable, and consists of the conjunction of sentences, obtained from the super-class specification and from each slot specification. In particular, for the super-classes $F_1, \ldots, F_h$ we have:

$$F_1(x) \wedge \ldots \wedge F_h(x)$$

and for a slot specification

$$
\begin{aligned}
&\textbf{MemberSlot}: \ S \\
&\textbf{ValueClass}: \ H \\
&\textbf{Cardinality.Min}: \ m \\
&\textbf{Cardinality.Max}: \ n
\end{aligned}
$$

we have

$$
\begin{aligned}
&\forall y. \left( S(x, y) \rightarrow H(y) \right) \ \wedge \\
&\exists y_1, \ldots, y_m. \left( \left( \bigwedge_{i \neq j} y_i \neq y_j \right) \wedge S(x, y_1) \wedge \cdots \wedge S(x, y_m) \right) \ \wedge \\
&\forall y_1, \ldots, y_{n+1}. \left( \left( S(x, y_1) \wedge \cdots \wedge S(x, y_{n+1}) \right) \rightarrow \bigvee_{i \neq j} y_i = y_j \right),
\end{aligned}
$$

under the assumption that within one frame definition the occurrences of $x$ refer to the same free variable. Finally the constraints on the slots are interpreted as conjunction, disjunction and negation, respectively, i.e.:

| | | |
|---|---|---|
| (INTERSECTION $H_1$ $H_2$) | is interpreted as | $H_1(x) \wedge H_2(x)$ |
| (UNION $H_1$ $H_2$) | is interpreted as | $H_1(x) \vee H_2(x)$ |
| (NOT $H$) | is interpreted as | $\neg H(x)$ |

A frame definition

$$\textbf{Frame}: \ F \ \textbf{in KB} \ \mathcal{F} \quad E$$

is then considered as the universally quantified sentence of the form

$$\forall x. (F(x) \rightarrow E(x)).$$

The whole frame knowledge base $\mathcal{F}$ is considered as the conjunction of all first-order sentences corresponding to the frame definitions in $\mathcal{F}$.

Here we regard frame definitions as necessary conditions, which is commonplace in the frame systems known as *assertional* frame systems, as opposed to *definitional* systems, where frame definitions are interpreted as necessary and sufficient conditions.

In order to enable the comparison with our formalisms for representing structured knowledge we restrict our attention to the reasoning tasks that involve the frame knowledge base, independently of the assertional knowledge, i.e., the frames instances. Fikes and Kehler (1985) mention several reasoning services associated with frames, such as:

- *Consistency checking*, which amounts to verifying whether a frame $F$ is satisfiable in a knowledge base. In particular, this involves both reasoning on cardinalities and checking whether the filler of a given slot belongs to a certain frame.

- *Inheritance*, which, in our case, amounts to the ability of identifying which of the frames are more general than a given frame, sometimes called *all-super-of* (Karp et al., 1995). All the properties of the more general frames are then inherited by the more specific one. Such a reasoning is therefore based on the more general ability to check the mutual relationhips between frame descriptions in the knowledge base.

These reasoning services are formalized in the first-order semantics as follows.

**Definition 3.3** Let $\mathcal{F}$ be a frame knowledge base and $F$ a frame in $\mathcal{F}$. We say that $F$ is *consistent in* $\mathcal{F}$ if the first-order sentence $\mathcal{F} \wedge \exists x.F(x)$ is satisfiable. Moreover, we say that a frame description $E$ is *more general* than $F$ in $\mathcal{F}$ if $\mathcal{F} \models \forall x.(F(x) \rightarrow E(x))$. ∎

## 3.3 Relationship between Frame Based Systems and ALUNI

The first-order semantics given above allows us to establish a straightforward relationship between frame languages and ALUNI. Indeed, we now present a translation from frame knowledge bases to ALUNI knowledge bases.

We first define the function $\theta$ that maps each frame expression into an $\mathcal{ALUNI}$ concept expression as follows:

- Every frame name $F$ is mapped into an atomic concept $\theta(F)$.

- Every slot name $S$ is mapped into an atomic role $\theta(S)$.

- Every slot constraint is mapped as follows

| | | |
|---|---|---|
| (UNION $H_1$ $H_2$) | is mapped into | $\theta(H_1) \sqcup \theta(H_2)$. |
| (INTERSECTION $H_1$ $H_2$) | is mapped into | $\theta(H_1) \sqcap \theta(H_2)$. |
| (NOT $H$) | is mapped into | $\neg\theta(H)$. |

- Every frame expression of the form

**SuperClasses** : $F_1, \ldots, F_h$
**MemberSlot** : $S_1$
  **ValueClass** : $H_1$
  **Cardinality.Min** : $m_1$
  **Cardinality.Max** : $n_1$
$\ldots$

**MemberSlot** : $S_k$
  **ValueClass** : $H_k$
  **Cardinality.Min** : $m_k$
  **Cardinality.Max** : $n_k$

$\mathcal{K} = (\mathcal{A}, \mathcal{P}, \mathcal{T})$, where

$\mathcal{A} = \{\texttt{Course}, \texttt{AdvCourse}, \texttt{BasCourse}, \texttt{Professor}, \texttt{Student}, \texttt{GradStudent}, \texttt{Undergrad}, \texttt{String}\}$,

$\mathcal{P} = \{\texttt{enrolls}, \texttt{taughtby}, \texttt{degree}\}$,

and the set $\mathcal{T}$ of assertions consists of:

$$\texttt{Course} \;\dot{\preceq}\; \forall\texttt{enrolls.Student} \sqcap \exists^{\geq 2}\texttt{enrolls} \sqcap \exists^{\leq 30}\texttt{enrolls} \sqcap$$
$$\forall\texttt{taughtby.}(\texttt{Professor} \sqcup \texttt{GradStudent}) \sqcap \exists^{=1}\texttt{taughtby}$$

$$\texttt{AdvCourse} \;\dot{\preceq}\; \texttt{Course} \sqcap \forall\texttt{enrolls.}(\texttt{GradStudent} \sqcap \neg\texttt{Undergrad}) \sqcap \exists^{\leq 20}\texttt{enrolls}$$

$$\texttt{BasCourse} \;\dot{\preceq}\; \texttt{Course} \sqcap \forall\texttt{taughtby.Professor}$$

$$\texttt{GradStudent} \;\dot{\preceq}\; \texttt{Student} \sqcap \forall\texttt{degree.String} \sqcap \exists^{=1}\texttt{degree}$$

$$\texttt{Undergrad} \;\dot{\preceq}\; \texttt{Student}$$

Figure 3: The ALUNI knowledge base corresponding to the KEE knowledge base in Figure 2

is mapped into the class expression

$$\theta(F_1) \sqcap \cdots \sqcap \theta(F_h) \sqcap$$
$$\forall\theta(S_1).\theta(H_1) \sqcap \exists^{\geq m_1}\theta(S_1) \sqcap \exists^{\leq n_1}\theta(S_1) \sqcap$$
$$\cdots$$
$$\forall\theta(S_k).\theta(H_k) \sqcap \exists^{\geq m_k}\theta(S_k) \sqcap \exists^{\leq n_k}\theta(S_k).$$

This mapping allows us to translate a frame knowledge base into an ALUNI knowledge base, as specified in the following definition.

**Definition 3.4** The ALUNI knowledge base $\theta(\mathcal{F}) = (\mathcal{A}, \mathcal{P}, \mathcal{T})$ corresponding to a frame knowledge base $\mathcal{F}$ is obtained as follows:

- $\mathcal{A}$ consists of one atomic concept $\theta(F)$ for each frame name $F$ in $\mathcal{F}$.

- $\mathcal{P}$ consists of one atomic role $\theta(S)$ for each slot name $S$ in $\mathcal{F}$.

- $\mathcal{T}$ consists of an inclusion assertion

$$\theta(F) \;\dot{\preceq}\; \theta(E)$$

for each frame definition

$$\textbf{Frame}: F \textbf{ in KB } \mathcal{F} \quad E$$

in $\mathcal{F}$. ∎

**Example 3.2 (cont.)** We illustrate the translation on the frame knowledge base in Figure 2. The corresponding ALUNI knowledge base is shown in Figure 3. ∎

The correctness of the translation follows from the correspondence between the set-theoretic semantics of ALUNI and the first-order interpretation of frames (see for example Hayes, 1979; Borgida, 1996; Donini et al., 1996). We can observe that inverse roles are in fact not necessary for the formalization of frames. Indeed, the possibility of referring to the inverse of a slot has been rarely considered in frame knowledge representation systems (Some exceptions are reported in Karp, 1992). Due to the absence of inverse roles the distinction between reasoning in finite and unrestricted models is not necessary[3]. Consequently, all the above mentioned forms of reasoning are captured by unrestricted concept consistency and concept subsumption in ALUNI knowledge bases. This is summarized in the following theorem.

**Theorem 3.5** *Let $\mathcal{F}$ be a frame knowledge-base, $F$ be a frame in $\mathcal{F}$, $E$ be a frame description, and $\theta(\mathcal{F})$, $\theta(F)$, and $\theta(E)$ be their translations in* ALUNI. *Then the following hold:*

- *$F$ is consistent in $\mathcal{F}$ if and only if $\theta(\mathcal{F}) \not\models_u \theta(F) \preceq \bot$.*

- *$E$ is more general than $F$ in $\mathcal{F}$ if and only if $\theta(\mathcal{F}) \models_u \theta(F) \preceq \theta(E)$.*

*Proof.* The claim directly follows from the semantics of frame knowledge bases and the translation into DLs that we have adopted. □

By Theorem 3.5 it becomes possible to exploit the methods for unrestricted reasoning on ALUNI knowledge bases in order to reason on frame knowledge bases. Since the problem of reasoning, e.g., in KEE is already EXPTIME-complete, we do not pay in terms of computational complexity for the expressiveness added by the constructs of ALUNI. In fact, by resorting to the correspondence with ALUNI it becomes possible to add to frame systems useful features, such as the possibility of specifying the inverse of a slot (Karp, 1992), and still retain the ability to reason in EXPTIME.

## 4. Semantic Data Models

Semantic data models were introduced primarily as formalisms for database schema design. They provide a means to model databases in a much richer way than traditional data models supported by Database Management Systems, and are becoming more and more important because they are adopted in most of the recent database design methodologies and Computer Aided Software Engineering tools.

The most widespread semantic data model is the Entity-Relationship (ER) model introduced by Chen (1976). It has by now become a standard, extensively used in the design phase of commercial applications. In the commonly accepted ER notation, classes are called *entities* and are represented as boxes, whereas relationships between entities are represented as diamonds. Arrows between entities, called *ISA* relationships, represent inclusion assertions. The links between entities and relationships represent the ER-*roles*, to which number restrictions are associated. Dashed links are used whenever such restrictions are refined for more specific entities. Finally, elementary properties of entities are modeled by *attributes*, whose values belong to one of several predefined domains, such as `Integer`, `String`, or `Boolean`.

---

3. If we eliminate from $\mathcal{ALUNI}$ inverse roles, then the resulting DL has the finite model property.

The ER model does not provide constructs for expressing explicit disjointness or disjunction of entities, but extensions of the model allow for the use of generalization hierarchies which represent a combination of these two constructs. In order to keep the presentation simple, we do not consider generalization hierarchies in the formalization we provide, although their addition would be straightforward. Similarly, we omit attributes of relations.

We now show that all relevant aspects of the ER model can be captured in ALUNI, and thus that reasoning on an ER schema can be reduced to reasoning on the corresponding ALUNI knowledge base. Since ALUNI is equipped with capabilities to reason on knowledge bases, both with respect to finite and unrestricted models (see Section 6.2), the reduction shows that reasoning on the ER model, and more generally on semantic data models, is decidable.

As in the case of frame-based systems, we restrict our attention to those aspects that constitute the core of the ER model. For this reason we do not consider some features, such as keys and weak entities, that have been introduced in the literature (Chen, 1976), but appear only in some of the formalizations of the ER model and the methodologies for conceptual modeling based on the model. A proposal for the treatment of keys in description logics is presented by Borgida and Weddell (1997).

In order to establish the correspondence between the ER model and ALUNI, we present formal syntax and semantics of ER schemata.

## 4.1 Syntax of the Entity-Relationship Model

Although the ER model has by now become an industrial standard, several variants and extensions have been introduced, which differ in minor aspects in expressiveness and in notation (Chen, 1976; Teorey, 1989; Batini, Ceri, & Navathe, 1992; Thalheim, 1992, 1993). Also, ER schemata are usually defined using a graphical notation which is particularly useful for an easy visualization of the data dependencies, but which is not well suited for our purposes. Therefore we have chosen a formalization of the ER model which abstracts with respect to the most important characteristics and allows us to develop the correspondence to ALUNI.

In the following, for two finite sets $X$ and $Y$ we call a function from a subset of $X$ to $Y$ an $X$-*labeled tuple over* $Y$. The labeled tuple $T$ that maps $x_i \in X$ to $y_i \in Y$, for $i \in \{1, \ldots, k\}$, is denoted $[x_1 : y_1, \ldots, x_k : y_k]$. We also write $T[x_i]$ to denote $y_i$.

**Definition 4.1** An *ER schema* is a tuple $\mathcal{S} = (\mathcal{L}_\mathcal{S}, \preceq_\mathcal{S}, att_\mathcal{S}, rel_\mathcal{S}, card_\mathcal{S})$, where

- $\mathcal{L}_\mathcal{S}$ is a finite alphabet partitioned into a set $\mathcal{E}_\mathcal{S}$ of *entity* symbols, a set $\mathcal{A}_\mathcal{S}$ of *attribute* symbols, a set $\mathcal{U}_\mathcal{S}$ of *role* symbols, a set $\mathcal{R}_\mathcal{S}$ of *relationship* symbols, and a set $\mathcal{D}_\mathcal{S}$ of *domain* symbols; each domain symbol $D$ has an associated predefined basic domain $D^{\mathcal{B}_\mathcal{D}}$, and we assume the various basic domains to be pairwise disjoint.

- $\preceq_\mathcal{S} \subseteq \mathcal{E}_\mathcal{S} \times \mathcal{E}_\mathcal{S}$ is a binary relation over $\mathcal{E}_\mathcal{S}$.

- $att_\mathcal{S}$ is a function that maps each entity symbol in $\mathcal{E}_\mathcal{S}$ to an $\mathcal{A}_\mathcal{S}$-labeled tuple over $\mathcal{D}_\mathcal{S}$.

- $rel_{\mathcal{S}}$ is a function that maps each relationship symbol in $\mathcal{R}_{\mathcal{S}}$ to an $\mathcal{U}_{\mathcal{S}}$-labeled tuple over $\mathcal{E}_S$. We assume without loss of generality that:

  - Each role is specific to exactly one relationship, i.e., for two relationships $R, R' \in \mathcal{R}_{\mathcal{S}}$ with $R \neq R'$, if $rel_{\mathcal{S}}(R) = [U_1 \colon E_1, \ldots, U_k \colon E_k]$ and $rel_{\mathcal{S}}(R') = [U'_1 \colon E'_1, \ldots, U'_{k'} \colon E'_{k'}]$, then $\{U_1, \ldots, U_k\}$ and $\{U'_1, \ldots, U'_{k'}\}$ are disjoint.

  - For each role $U \in \mathcal{U}_{\mathcal{S}}$ there is a relationship $R$ and an entity $E$ such that $rel_{\mathcal{S}}(R) = [\ldots, U \colon E, \ldots]$.

- $card_{\mathcal{S}}$ is a function from $\mathcal{E}_{\mathcal{S}} \times \mathcal{R}_{\mathcal{S}} \times \mathcal{U}_{\mathcal{S}}$ to $\mathbb{N}_0 \times (\mathbb{N}_0 \cup \{\infty\})$ that satisfies the following condition: for a relationship $R \in \mathcal{R}_{\mathcal{S}}$ such that $rel_{\mathcal{S}}(R) = [U_1 \colon E_1, \ldots, U_k \colon E_k]$, $card_{\mathcal{S}}(E, R, U)$ is defined only if $U = U_i$ for some $i \in \{1, \ldots, k\}$, and if $E \preceq^*_{\mathcal{S}} E_i$ (where $\preceq^*_{\mathcal{S}}$ denotes the reflexive transitive closure of $\preceq_{\mathcal{S}}$). The first component of $card_{\mathcal{S}}(E, R, U)$ is denoted with $cmin_{\mathcal{S}}(E, R, U)$ and the second component with $cmax_{\mathcal{S}}(E, R, U)$. If not stated otherwise, $cmin_{\mathcal{S}}(E, R, U)$ is assumed to be 0 and $cmax_{\mathcal{S}}(E, R, U)$ is assumed to be $\infty$. ∎

Before specifying the formal semantics of ER schemata we give an intuitive description of the components of a schema. The relation $\preceq_{\mathcal{S}}$ models the ISA-relationship between entities. We do not need to make any special assumption on the form of $\preceq_{\mathcal{S}}$ such as acyclicity or injectivity. The function $att_{\mathcal{S}}$ is used to model attributes of entities. If for example $att_{\mathcal{S}}$ associates the $\mathcal{A}_{\mathcal{S}}$-labeled tuple $[A_1 \colon \texttt{Integer}, A_2 \colon \texttt{String}]$ to an entity $E$, then $E$ has two attributes $A_1, A_2$ whose values are integers and strings respectively. For simplicity we assume attributes to be single-valued and mandatory, but we could easily handle also multi-valued attributes with associated cardinalities. The function $rel_{\mathcal{S}}$ associates a set of roles to each relationship symbol $R$, determining implicitly also the arity of $R$, and for each role $U$ in such set a distinguished entity, called the *primary entity for $U$ in $R$*. In a database satisfying the schema only instances of the primary entity are allowed to participate in the relationship via the role $U$. The function $card_{\mathcal{S}}$ specifies *cardinality constraints*, i.e., constraints on the minimum and maximum number of times an instance of an entity may participate in a relationship via some role. Since such constraints are meaningful only if the entity can effectively participate in the relationship, the function is defined only for the sub-entities of the primary entity. The special value $\infty$ is used when no restriction is posed on the maximum cardinality. Such constraints can be used to specify both existence dependencies and functionality of relations (Cosmadakis & Kanellakis, 1986). They are often used only in a restricted form, where the minimum cardinality is either 0 or 1 and the maximum cardinality is either 1 or $\infty$. Cardinality constraints in the form considered here have been introduced already by Abrial (1974) and subsequently studied by Grant and Minker (1984), Lenzerini and Nobili (1990), Ferg (1991), Ye, Parent, and Spaccapietra (1994), Thalheim (1992).

**Example 4.2** Figure 4 shows a simple ER schema modeling a state of affairs similar to the one represented by the KEE knowledge base in Figure 2. We have used the standard graphic notation for ER schemata, except for the dashed link, which represents the refinement of a cardinality constraint for the participation of a sub-entity (in our case `AdvCourse`) in a relationship (in our case `ENROLLING`). ∎

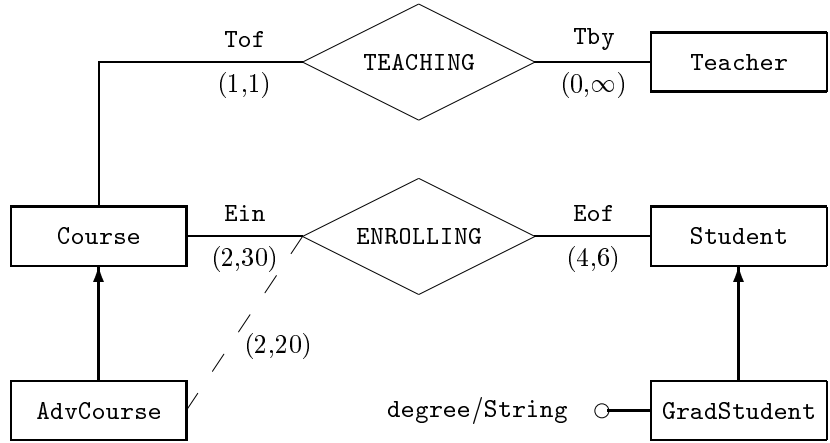

Figure 4: An ER schema

## 4.2 Semantics of the Entity-Relationship Model

The semantics of an ER schema can be given by specifying which database states are consistent with the information structure represented by the schema. Formally, a database state $\mathcal{B}$ corresponding to an ER schema $\mathcal{S} = (\mathcal{L}_\mathcal{S}, \preceq_\mathcal{S}, att_\mathcal{S}, rel_\mathcal{S}, card_\mathcal{S})$ is constituted by a nonempty *finite* set $\Delta^\mathcal{B}$, assumed to be disjoint from all basic domains, and a function $\cdot^\mathcal{B}$ that maps

- every domain symbol $D \in \mathcal{D}_\mathcal{S}$ to the corresponding basic domain $D^{\mathcal{B}_\mathcal{D}}$,
- every entity $E \in \mathcal{E}_\mathcal{S}$ to a subset $E^\mathcal{B}$ of $\Delta^\mathcal{B}$,
- every attribute $A \in \mathcal{A}_\mathcal{S}$ to a set $A^\mathcal{B} \subseteq \Delta^\mathcal{B} \times \bigcup_{D \in \mathcal{D}_\mathcal{S}} D^{\mathcal{B}_\mathcal{D}}$, and
- every relationship $R \in \mathcal{R}_\mathcal{S}$ to a set $R^\mathcal{B}$ of $\mathcal{U}_\mathcal{S}$-labeled tuples over $\Delta^\mathcal{B}$.

The elements of $E^\mathcal{B}$, $A^\mathcal{B}$, and $R^\mathcal{B}$ are called *instances* of $E$, $A$, and $R$ respectively.

A database state is considered acceptable if it satisfies all integrity constraints that are part of the schema. This is captured by the definition of legal database state.

**Definition 4.3** A database state $\mathcal{B}$ is said to be *legal for* an ER schema $\mathcal{S} = (\mathcal{L}_\mathcal{S}, \preceq_\mathcal{S}, att_\mathcal{S}, rel_\mathcal{S}, card_\mathcal{S})$, if it satisfies the following conditions:

- For each pair of entities $E_1, E_2 \in \mathcal{E}_\mathcal{S}$ such that $E_1 \preceq_\mathcal{S} E_2$, it holds that $E_1^\mathcal{B} \subseteq E_2^\mathcal{B}$.

- For each entity $E \in \mathcal{E}_\mathcal{S}$, if $att_\mathcal{S}(E) = [A_1 : D_1, \ldots, A_h : D_h]$, then for each instance $e \in E^\mathcal{B}$ and for each $i \in \{1, \ldots, h\}$ the following holds:

  - there is exactly one element $a_i \in A_i^\mathcal{B}$ whose first component is $e$, and
  - the second component of $a_i$ is an element of $D_i^{\mathcal{B}_\mathcal{D}}$.

- For each relationship $R \in \mathcal{R}_\mathcal{S}$ such that $rel_\mathcal{S}(R) = [U_1 : E_1, \ldots, U_k : E_k]$, all instances of $R$ are of the form $[U_1 : e_1, \ldots, U_k : e_k]$, where $e_i \in E_i^\mathcal{B}$, $i \in \{1, \ldots, k\}$.

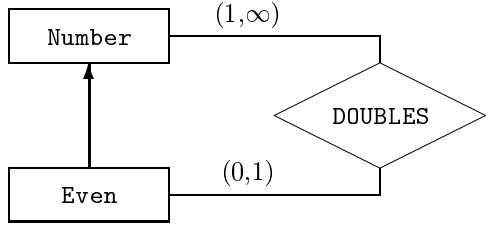

Figure 5: The ER schema corresponding to Example 2.2

- For each relationship $R \in \mathcal{R}_{\mathcal{S}}$ such that $rel_{\mathcal{S}}(R) = [U_1:E_1, \ldots, U_k:E_k]$, for each $i \in \{1, \ldots, k\}$, for each entity $E \in \mathcal{E}_{\mathcal{S}}$ such that $E \preceq^*_{\mathcal{S}} E_i$ and for each instance $e$ of $E$ in $\mathcal{I}$, it holds that

$$cmin_{\mathcal{S}}(E, R, U_i) \leq \sharp\{r \in R^{\mathcal{B}} \mid r[U_i] = e\} \leq cmax_{\mathcal{S}}(E, R, U_i).$$

∎

Notice that the definition of database state reflects the usual assumption in databases that database states are finite structures (see also Cosmadakis, Kanellakis, & Vardi, 1990). In fact, the basic domains are not required to be finite, but for each legal database state for a schema, only a finite set of values from the basic domains are actually of interest. We define the *active domain* $\Delta^{\mathcal{B}}_{act}$ of a database state $\mathcal{B}$ as the set of all elements of the basic domains $D^{\mathcal{B}_{\mathcal{D}}}$, $D \in \mathcal{D}_{\mathcal{S}}$, that effectively appear as values of attributes in $\mathcal{B}$. More formally:

$$\Delta^{\mathcal{B}}_{act} = \{d \in D^{\mathcal{B}_{\mathcal{D}}} \mid D \in \mathcal{D}_{\mathcal{S}} \wedge \exists A \in \mathcal{A}_{\mathcal{S}}, e \in \Delta^{\mathcal{B}} . (e, d) \in A^{\mathcal{B}}\}.$$

Since $\Delta^{\mathcal{B}}$ is finite and $\mathcal{A}_{\mathcal{S}}$ contains only a finite number of attributes, which are functional by definition, also $\Delta^{\mathcal{B}}_{act}$ is finite.

Reasoning in the ER model includes verifying entity satisfiability and deducing inheritance. *Entity satisfiability* amounts to checking whether a given entity can be populated in some legal database state (Atzeni & Parker Jr., 1986; Lenzerini & Nobili, 1990; Di Battista & Lenzerini, 1993), and corresponds to the notion of concept consistency in DLs. Deducing *inheritance* amounts to verifying whether in all databases that are legal for the schema, every instance of an entity is also an instance of another entity. Such implied ISA relationships can arise for different reasons. Either trivially, through the transitive closure of the explicit ISA relationships present in the schema, or in more subtle ways, through specific patterns of cardinality restrictions along cycles in the schema and the requirement of the database state to be finite (Lenzerini & Nobili, 1990; Cosmadakis et al., 1990).

**Example 4.4** Figure 5 shows an ER schema modeling the same situation as the knowledge base of Example 2.2. Arguing exactly as in that example we can conclude that the two entities `Number` and `Even` denote the same set of elements in every finite database legal for the schema, although the ISA relation from `Number` to `Even` is not stated explicitly. It is implied, however, due to the cycle involving the relationship and the two entities and due to the particular form of cardinality constraints. ∎

### 4.3 Relationship between Entity-Relationship Schemata and ALUNI

We now show that the different forms of reasoning on ER schemata are captured by finite concept consistency and finite concept subsumption in ALUNI. The correspondence between the two formalisms is established by defining a translation $\phi$ from ER schemata to ALUNI knowledge bases, and then proving that there is a correspondence between legal database states and finite models of the derived knowledge base.

**Definition 4.5** Let $\mathcal{S} = (\mathcal{L}_\mathcal{S}, \preceq_\mathcal{S}, att_\mathcal{S}, rel_\mathcal{S}, card_\mathcal{S})$ be an ER schema. The ALUNI knowledge base $\phi(\mathcal{S}) = (\mathcal{A}, \mathcal{P}, \mathcal{T})$ is defined as follows:

The set $\mathcal{A}$ of atomic concepts of $\phi(\mathcal{S})$ contains the following elements:

- for each domain symbol $D \in \mathcal{D}_\mathcal{S}$, an atomic concept $\phi(D)$;
- for each entity $E \in \mathcal{E}_\mathcal{S}$, an atomic concept $\phi(E)$;
- for each relationship $R \in \mathcal{R}_\mathcal{S}$, an atomic concept $\phi(R)$.

The set $\mathcal{P}$ of atomic roles of $\phi(\mathcal{S})$ contains the following elements:

- for each attribute $A \in \mathcal{A}_\mathcal{S}$, an atomic role $\phi(A)$;
- for each relationship $R \in \mathcal{R}_\mathcal{S}$ such that $rel_\mathcal{S}(R) = [U_1:E_1, \ldots, U_k:E_k]$, $k$ atomic roles $\phi(U_1), \ldots, \phi(U_k)$.

The set $\mathcal{T}$ of assertions of $\phi(\mathcal{S})$ contains the following elements:

- for each pair of entities $E_1, E_2 \in \mathcal{E}_\mathcal{S}$ such that $E_1 \preceq_\mathcal{S} E_2$, the assertion

$$\phi(E_1) \; \dot{\preceq} \; \phi(E_2) \tag{1}$$

- for each entity $E \in \mathcal{E}_\mathcal{S}$ such that $att_\mathcal{S}(E) = [A_1:D_1, \ldots, A_h:D_h]$, the assertion

$$\phi(E) \; \dot{\preceq} \; \forall \phi(A_1).\phi(D_1) \sqcap \cdots \sqcap \forall \phi(A_h).\phi(D_h) \sqcap \exists^{=1}\phi(A_1) \sqcap \cdots \sqcap \exists^{=1}\phi(A_h) \tag{2}$$

- for each relationship $R \in \mathcal{R}_\mathcal{S}$ such that $rel_\mathcal{S}(R) = [U_1:E_1, \ldots, U_k:E_k]$, the assertions

$$\phi(R) \; \dot{\preceq} \; \forall \phi(U_1).\phi(E_1) \sqcap \cdots \sqcap \forall \phi(U_k).\phi(E_k) \sqcap \exists^{=1}\phi(U_1) \sqcap \cdots \sqcap \exists^{=1}\phi(U_k) \tag{3}$$

$$\phi(E_i) \; \dot{\preceq} \; \forall (\phi(U_i))^-.\phi(R), \qquad i \in \{1, \ldots, k\} \tag{4}$$

- for each relationship $R \in \mathcal{R}_\mathcal{S}$ such that $rel_\mathcal{S}(R) = [U_1:E_1, \ldots, U_k:E_k]$, for $i \in \{1, \ldots, k\}$, and for each entity $E \in \mathcal{E}_\mathcal{S}$ such that $E \preceq_\mathcal{S}^* E_i$,

  - if $m = cmin_\mathcal{S}(E, R, U_i) \neq 0$, the assertion

$$\phi(E) \; \dot{\preceq} \; \exists^{\geq m}(\phi(U_i))^-. \tag{5}$$

  - if $n = cmax_\mathcal{S}(E, R, U_i) \neq \infty$, the assertion

$$\phi(E) \; \dot{\preceq} \; \exists^{\leq n}(\phi(U_i))^-. \tag{6}$$

- for each pair of symbols $X_1, X_2 \in \mathcal{E}_\mathcal{S} \cup \mathcal{R}_\mathcal{S} \cup \mathcal{D}_\mathcal{S}$ such that $X_1 \neq X_2$ and $X_1 \in \mathcal{R}_\mathcal{S} \cup \mathcal{D}_\mathcal{S}$, the assertion

$$\phi(X_1) \; \dot{\preceq} \; \neg\phi(X_2). \tag{7}$$

∎

$\mathcal{K} = (\mathcal{A}, \mathcal{P}, \mathcal{T})$, where

$\mathcal{A} = \{\texttt{Course}, \texttt{AdvCourse}, \texttt{Teacher}, \texttt{Student}, \texttt{GradStudent}, \texttt{TEACHING}, \texttt{ENROLLING}, \texttt{String}\}$,

$\mathcal{P} = \{\texttt{Tof}, \texttt{Tby}, \texttt{Ein}, \texttt{Eof}, \texttt{degree}\}$,

and the set $\mathcal{T}$ of assertions consists of:

$$
\begin{aligned}
\texttt{TEACHING} &\;\stackrel{\cdot}{\preceq}\; \forall\texttt{Tof}.\texttt{Course} \sqcap \exists^{=1}\texttt{Tof} \sqcap \\
&\qquad \forall\texttt{Tby}.\texttt{Teacher} \sqcap \exists^{=1}\texttt{Tby} \\
\texttt{ENROLLING} &\;\stackrel{\cdot}{\preceq}\; \forall\texttt{Ein}.\texttt{Course} \sqcap \exists^{=1}\texttt{Ein} \sqcap \\
&\qquad \forall\texttt{Eof}.\texttt{Student} \sqcap \exists^{=1}\texttt{Eof} \\
\texttt{Course} &\;\stackrel{\cdot}{\preceq}\; \forall\texttt{Tof}^-.\texttt{TEACHING} \sqcap \exists^{=1}\texttt{Tof}^- \sqcap \\
&\qquad \forall\texttt{Ein}^-.\texttt{ENROLLING} \sqcap \exists^{\geq 2}\texttt{Ein}^- \sqcap \exists^{\leq 30}\texttt{Ein}^- \\
\texttt{AdvCourse} &\;\stackrel{\cdot}{\preceq}\; \texttt{Course} \sqcap \exists^{\leq 20}\texttt{Ein}^- \\
\texttt{Teacher} &\;\stackrel{\cdot}{\preceq}\; \forall\texttt{Tby}^-.\texttt{TEACHING} \\
\texttt{Student} &\;\stackrel{\cdot}{\preceq}\; \forall\texttt{Eof}^-.\texttt{ENROLLING} \sqcap \exists^{\geq 4}\texttt{Eof}^- \sqcap \exists^{\leq 6}\texttt{Eof}^- \\
\texttt{GradStudent} &\;\stackrel{\cdot}{\preceq}\; \texttt{Student} \sqcap \forall\texttt{degree}.\texttt{String} \sqcap \exists^{=1}\texttt{degree}.
\end{aligned}
$$

Figure 6: The ALUNI knowledge base corresponding to the ER schema in Figure 4

**Example 4.2 (cont.)** We illustrate the translation on the ER schema of Figure 4. The ALUNI knowledge base that captures exactly its semantics is shown in Figure 6, where for brevity the disjointness assertions (7) are omitted, and assertions with the same concept on the left hand side are collapsed. ∎

The translation makes use of both inverse attributes and number restrictions to capture the semantics of ER schemata. We observe that, by means of the inverse constructor, a binary relationship could be treated in a simpler way by choosing a traversal direction and mapping the relationship directly to a role. Notice also that the assumption of acyclicity of the resulting knowledge base is unrealistic in this case, and in order to exploit the correspondence for reasoning in the ER model, we need techniques that can deal with inverse attributes, number restrictions, and cycles together. As shown in Example 2.2, the combination of these factors causes the finite model property to fail to hold, and we need to resort to reasoning methods for finite models.

In fact, we can reduce reasoning in the ER model to finite model reasoning in ALUNI knowledge bases. For this purpose we define a mapping between database states corresponding to an ER schema and finite interpretations of the knowledge base derived from it. Due to the possible presence of relations with arity greater than 2, this mapping is however not one-to-one and we first need to characterize those interpretations of the knowledge base that directly correspond to database states.

**Definition 4.6** Let $\mathcal{S} = (\mathcal{L}_\mathcal{S}, \preceq_\mathcal{S}, att_\mathcal{S}, rel_\mathcal{S}, card_\mathcal{S})$ be an ER schema and $\phi(\mathcal{S})$ be defined as above. An interpretation $\mathcal{I}$ of $\phi(\mathcal{S})$ is *relation-descriptive*, if for every relationship $R \in \mathcal{R}_\mathcal{S}$, with $rel_\mathcal{S}(R) = [U_1:E_1, \ldots, U_k:E_k]$, for every $d, d' \in (\phi(R))^\mathcal{I}$, we have that

$$
(\bigwedge_{1 \leq i \leq k} \forall d'' \in \Delta^\mathcal{I} . ((d, d'') \in (\phi(U_i))^\mathcal{I} \leftrightarrow (d', d'') \in (\phi(U_i))^\mathcal{I})) \rightarrow d = d'. \tag{8}
$$

∎

Intuitively, the extension of a relationship in a database state is a *set* of labeled tuples, and such a set does not contain the same element twice. Therefore it is implicit in the semantics of the ER model that there cannot be two labeled tuples connected through all roles of the relationship to exactly the same elements of the domain. In a model of the ALUNI knowledge base corresponding to the ER schema, on the other hand, each tuple is represented by a new individual, and the above condition is not implicit anymore. It also cannot be imposed in ALUNI by suitable assertions. The following lemma, however, shows that we do not need such an explicit condition, when we are interested in reasoning on an ALUNI knowledge base corresponding to an ER schema. This is due to the fact that we can always restrict ourselves to considering only relation-descriptive models.

**Lemma 4.7** *Let $\mathcal{S}$ be an ER schema, $\phi(\mathcal{S})$ be the ALUNI knowledge base obtained from $\mathcal{S}$ according to Definition 4.5, and $C$ be a concept expression of $\phi(\mathcal{S})$. If $C$ is finitely consistent in $\phi(\mathcal{S})$, then there is a finite relation-descriptive model $\mathcal{I}$ of $\phi(\mathcal{S})$ such that $C^{\mathcal{I}} \neq \emptyset$.*

*Proof.* Let $\mathcal{I}_0$ be a finite model of $\phi(\mathcal{S})$ such that $C^{\mathcal{I}} \neq \emptyset$. We can build a finite relation-descriptive model $\mathcal{I}'$ by starting from $\mathcal{I}_0$ and applying the following construction once for each relationship in $\mathcal{R}_{\mathcal{S}}$.

Let $\mathcal{I}$ be the model obtained in the previous step and let $R \in \mathcal{R}_{\mathcal{S}}$ with $rel_{\mathcal{S}}(R) = [U_1 : E_1, \ldots, U_k : E_k]$ be the next relationship to which we apply the construction. We construct from $\mathcal{I}$ a model $\mathcal{I}_R$ such that condition 8 is satisfied for relationship $R$.

Given an individual $r \in (\phi(R))^{\mathcal{I}}$, we denote by $U_i(d)$, $i \in \{1, \ldots, k\}$ the (unique) individual $e$ such that $(r, e) \in (\phi(U_i))^{\mathcal{I}}$. For $e_i \in (\phi(E_i))^{\mathcal{I}}$, $i \in \{1, \ldots, k\}$ we define $X_{(U_1 : e_1, \ldots, U_k : e_k)} = \{r \in (\phi(R))^{\mathcal{I}} \mid U_i(d) = e_i, \text{ for } i \in \{1, \ldots, k\}\}$. We call *conflict-set* a set $X_{(U_1 : e_1, \ldots, U_k : e_k)}$ with more than one element. From each conflict-set $X_{(U_1 : e_1, \ldots, U_k : e_k)}$ we randomly choose one individual $r$, and we say that the others *induce a conflict* on $(U_1 : e_1, \ldots, U_k : e_k)$. We call *Conf* the (finite) set of all objects inducing a conflict on some $(U_1 : e_1, \ldots, U_k : e_k)$.

We define an interpretation $\mathcal{I}_{2^{Conf}}$ as the disjoint union of $2^{\sharp Conf}$ copies of $\mathcal{I}$, one copy, denoted by $\mathcal{I}_{\mathcal{Z}}$, for every set $\mathcal{Z} \in 2^{Conf}$. We denote by $d_{\mathcal{Z}}$ the copy in $\mathcal{I}_{\mathcal{Z}}$ of the individual $d$ in $\mathcal{I}$. Since the disjoint union of two models of an ALUNI knowledge base is again a model, $\mathcal{I}_{2^{Conf}}$ is a model of $\phi(\mathcal{S})$. Let $\mathcal{I}_{\mathcal{Z}}$ and $\mathcal{I}_{\mathcal{Z}'}$ be two copies of $\mathcal{I}$ in $\mathcal{I}_{2^{Conf}}$. We call *exchanging $U_k(r_{\mathcal{Z}})$ with $U_k(r_{\mathcal{Z}'})$* the operation on $\mathcal{I}_{2^{Conf}}$ consisting of replacing in $(\phi(U_k))^{\mathcal{I}_{\mathcal{Z}}}$ the pair $(r_{\mathcal{Z}}, U_k(r_{\mathcal{Z}}))$ with $(r_{\mathcal{Z}}, U_k(r_{\mathcal{Z}'}))$ and, at the same time, replacing in $(\phi(U_k))^{\mathcal{I}_{\mathcal{Z}'}}$ the pair $(r_{\mathcal{Z}'}, U_k(r_{\mathcal{Z}'}))$ with $(r_{\mathcal{Z}'}, U_k(r_{\mathcal{Z}}))$. Intuitively, by exchanging $U_k(r_{\mathcal{Z}})$ with $U_k(r_{\mathcal{Z}'})$, the individuals $r_{\mathcal{Z}}$ and $r_{\mathcal{Z}'}$ do not induce conflicts anymore.

We construct now from $\mathcal{I}_{2^{Conf}}$ an interpretation $\mathcal{I}_R$ as follows: For each $r \in Conf$ and for each $\mathcal{Z} \in 2^{Conf}$ such that $r \in \mathcal{Z}$, we exchange $U_k(r_{\mathcal{Z}})$ with $U_k(r_{\mathcal{Z} \setminus \{r\}})$. It is possible to show that all conflicts are thus eliminated while no new conflict is created. Hence, in $\mathcal{I}_R$, condition 8 for $R$ is satisfied. We still have to show that $\mathcal{I}_R$ is a model of $\phi(\mathcal{S})$ in which $C^{\mathcal{I}_R} \neq \emptyset$. Indeed, it is straightforward to check by induction that for every concept expression $C'$ appearing in $\phi(\mathcal{S})$, for all $\mathcal{Z} \in 2^{Conf}$, $d \in C'^{\mathcal{I}}$ if and only if $d_{\mathcal{Z}} \in C'^{\mathcal{I}_R}$. Thus all assertions of $\phi(\mathcal{S})$ are still satisfied in $\mathcal{I}_R$ and $C^{\mathcal{I}_R} \neq \emptyset$. □

With this result, the following correspondence between legal database states for an ER schema and relation-descriptive models of the resulting ALUNI knowledge base can be established.

**Proposition 4.8** *For every ER schema $\mathcal{S} = (\mathcal{L}_{\mathcal{S}}, \preceq_{\mathcal{S}}, att_{\mathcal{S}}, rel_{\mathcal{S}}, card_{\mathcal{S}})$ there exist two mappings $\alpha_{\mathcal{S}}$, from database states corresponding to $\mathcal{S}$ to finite interpretations of its translation $\phi(\mathcal{S})$, and $\beta_{\mathcal{S}}$, from finite relation-descriptive interpretations of $\phi(\mathcal{S})$ to database states corresponding to $\mathcal{S}$, such that:*

1. *For each legal database state $\mathcal{B}$ for $\mathcal{S}$, $\alpha_{\mathcal{S}}(\mathcal{B})$ is a finite model of $\phi(\mathcal{S})$, and for each symbol $X \in \mathcal{E}_{\mathcal{S}} \cup \mathcal{A}_{\mathcal{S}} \cup \mathcal{R}_{\mathcal{S}} \cup \mathcal{D}_{\mathcal{S}}$, $X^{\mathcal{B}} = (\phi(X))^{\alpha_{\mathcal{S}}(\mathcal{B})}$.*

2. *For each finite relation-descriptive model $\mathcal{I}$ of $\phi(\mathcal{S})$, $\beta_{\mathcal{S}}(\mathcal{I})$ is a legal database state for $\mathcal{S}$, for each entity $E \in \mathcal{E}_{\mathcal{S}}$, $(\phi(E))^{\mathcal{I}} = E^{\beta_{\mathcal{S}}(\mathcal{I})}$, and for each symbol $X \in \mathcal{A}_{\mathcal{S}} \cup \mathcal{R}_{\mathcal{S}} \cup \mathcal{D}_{\mathcal{S}}$, $\sharp\phi(X)^{\mathcal{I}} = \sharp X^{\beta_{\mathcal{S}}(\mathcal{I})}$.*

*Proof.* (1) Given a database state $\mathcal{B}$, we define the interpretation $\mathcal{I} = \alpha_{\mathcal{S}}(\mathcal{B})$ of $\phi(\mathcal{S})$ as follows:

- $\Delta^{\mathcal{I}} = \Delta^{\mathcal{B}} \cup \Delta^{\mathcal{B}}_{act} \cup \bigcup_{R \in \mathcal{R}_{\mathcal{S}}} R^{\mathcal{B}}$.

- For each symbol $X \in \mathcal{E}_{\mathcal{S}} \cup \mathcal{A}_{\mathcal{S}} \cup \mathcal{R}_{\mathcal{S}} \cup \mathcal{D}_{\mathcal{S}}$,

$$(\phi(X))^{\mathcal{I}} = X^{\mathcal{B}}. \tag{9}$$

- For each relationship $R \in \mathcal{R}_{\mathcal{S}}$ such that $rel_{\mathcal{S}}(R) = [U_1 : E_1, \ldots, U_k : E_k]$,

$$(\phi(U_i))^{\mathcal{I}} = \{(r, e) \in \Delta^{\mathcal{I}} \times \Delta^{\mathcal{I}} \mid r \in R^{\mathcal{B}}, \text{ and } r[U_i] = e\}, \quad i \in \{1, \ldots, k\}. \tag{10}$$

Let $\mathcal{B}$ be a legal database state. To prove claim (1) it is sufficient to show that $\mathcal{I}$ satisfies every assertion in $\phi(\mathcal{S})$. Assertions 1 are satisfied since $\mathcal{B}$ satisfies the set inclusion between the extensions of the corresponding entities. With respect to assertions 2, let $E \in \mathcal{E}_{\mathcal{S}}$ be an entity such that $att_{\mathcal{S}}(E) = [A_1 : D_1, \ldots, A_h : D_h]$, and consider an instance $e \in (\phi(E))^{\mathcal{I}}$. We have to show that for each $i \in \{1, \ldots, h\}$, there is exactly one element $e_i \in \Delta^{\mathcal{I}}$ such that $(e, e_i) \in (\phi(A_i))^{\mathcal{I}}$, and moreover that $e_i \in (\phi(D_i))^{\mathcal{I}}$. By 9, $e \in E^{\mathcal{B}}$, and by definition of legal database state there is exactly one element $a_i \in A_i^{\mathcal{B}} = (\phi(A_i))^{\mathcal{I}}$ whose first component is $e$. Moreover, the second component $e_i$ of $a_i$ is an element of $D_i^{\mathcal{B}_{\mathcal{D}}} = (\phi(D_i))^{\mathcal{I}}$. With respect to assertions 3, let $R \in \mathcal{R}_{\mathcal{S}}$ be a relationship such that $rel_{\mathcal{S}}(R) = [U_1 : E_1, \ldots, U_k : E_k]$, and consider an instance $r \in (\phi(R))^{\mathcal{I}}$. We have to show that for each $i \in \{1, \ldots, k\}$ there is exactly one element $e_i \in \Delta^{\mathcal{I}}$ such that $(r, e_i) \in (\phi(U_i))^{\mathcal{I}}$, and that moreover $e_i \in (\phi(E_i))^{\mathcal{I}}$. By 9, $r \in R^{\mathcal{B}}$, and by definition of legal database state, $r$ is a labeled tuple of the form $[U_1 : e'_1, \ldots, U_k : e'_k]$, where $e'_i \in E_i^{\mathcal{B}}$, $i \in \{1, \ldots, k\}$. Therefore $r$ is a function defined on $\{U_1, \ldots, U_k\}$, and by 10, $e_i$ is unique and equal to $e'_i$. Moreover, again by 9, $e_i \in (\phi(E_i))^{\mathcal{I}} = E_i^{\mathcal{B}}$. Assertions 4 are satisfied, since by 10 the first component of each element of $(\phi(U_i))^{\mathcal{I}}$ is always an element of $R^{\mathcal{B}} = (\phi(R))^{\mathcal{I}}$. With respect to assertions 5, let $R \in \mathcal{R}_{\mathcal{S}}$ be a relationship such that $rel_{\mathcal{S}}(R) = [U_1 : E_1, \ldots, U_k : E_k]$, let $E \in \mathcal{E}_{\mathcal{S}}$ be an entity such that $E \preceq_{\mathcal{S}} E_i$, for some $i \in \{1, \ldots, k\}$, and such that $m = cmin_{\mathcal{S}}(E, R, U_i) \neq 0$.

Consider an instance $e \in (\phi(E))^{\mathcal{I}}$. We have to show that there are at least $m$ pairs in $(\phi(U_i))^{\mathcal{I}}$ that have $e$ as their second component. Since assertions 4 are satisfied we know that the first component of all such pairs is an instance of $\phi(R)$. By 9 and by definition of legal database state, there are at least $m$ labeled tuples in $R^{\mathcal{B}}$ whose $U_i$ component is equal to $e$. By 10, $(\phi(U_i))^{\mathcal{I}}$ contains at least $m$ pairs whose second component is equal to $e$. With respect to assertions 6 we can proceed in a similar way. Finally, assertions 7 are satisfied since first, by definition the basic domains are pairwise disjoint and disjoint from $\Delta^{\mathcal{B}}$ and from the set of labeled tuples, second, no element of $\Delta^{\mathcal{B}}$ is a labeled tuple, and third, labeled tuples corresponding to different relationships cannot be equal since they are defined over different sets of roles.

(2) Let $\mathcal{I}$ be a finite relation-descriptive interpretation of $\phi(\mathcal{S})$. For each basic domain $D \in \mathcal{D}_{\mathcal{S}}$, let $\beta_{\Delta}^{D}$ be a function from $\Delta^{\mathcal{I}}$ to $D^{\mathcal{B}_{\mathcal{D}}}$ that is one-to-one and onto. Since $\Delta^{\mathcal{I}}$ is finite and each basic domain contains a countable number of elements, such a function always exists. In order to define $\beta_{\mathcal{S}}(\mathcal{I})$ we first specify a mapping $\beta_{\Delta}$ that associates to each individual $d \in \Delta^{\mathcal{I}}$ an element as follows:

- If $d \in (\phi(E))^{\mathcal{I}}$ for some entity $E \in \mathcal{E}_{\mathcal{S}}$, then $\beta_{\Delta}(d) = d$.

- If $d \in (\phi(R))^{\mathcal{I}}$ for some relationship $R \in \mathcal{R}_{\mathcal{S}}$ with $rel_{\mathcal{S}}(R) = [U_1 : E_1, \ldots, U_k : E_k]$, and there are individuals $d_1, \ldots, d_k \in \Delta^{\mathcal{I}}$ such that $(d, d_i) \in (\phi(U_i))^{\mathcal{I}}$, for $i \in \{1, \ldots, k\}$, then $\beta_{\Delta}(d) = [U_1 : d_1, \ldots, U_k : d_k]$.

- If $d \in (\phi(D))^{\mathcal{I}}$ for some basic domain $D \in \mathcal{D}_{\mathcal{S}}$, then $\beta_{\Delta}(d) = \beta_{\Delta}^{D}(d)$.

- Otherwise $\beta_{\Delta}(d) = d$.

For a pair of individuals $(d_1, d_2) \in \Delta^{\mathcal{I}} \times \Delta^{\mathcal{I}}$, $\beta_{\Delta}((d_1, d_2)) = (\beta_{\Delta}(d_1), \beta_{\Delta}(d_2))$, and for a set $X$, $\beta_{\Delta}(X) = \{\beta_{\Delta}(x) \mid x \in X\}$.

If $\mathcal{I}$ is a model of $\phi(\mathcal{S})$ the above rules define $\beta_{\Delta}(d)$ for every $d \in \Delta^{\mathcal{I}}$. Indeed, by assertions 7, each $d \in \Delta^{\mathcal{I}}$ can be an instance of at most one atomic concept corresponding to a relationship or basic domain, and if this is the case it is not an instance of any atomic concept corresponding to an entity. Moreover, if $d \in (\phi(R))^{\mathcal{I}}$ for some relationship $R \in \mathcal{R}_{\mathcal{S}}$ with $rel_{\mathcal{S}}(R) = [U_1 : E_1, \ldots, U_k : E_k]$, then by assertions 3, for each $i \in \{1, \ldots, k\}$ there is exactly one element $d_i \in \Delta^{\mathcal{I}}$ such that $(d, d_i) \in (\phi(U_i))^{\mathcal{I}}$. If $\mathcal{I}$ is not a model of $\phi(\mathcal{S})$ and for some $d \in \Delta^{\mathcal{I}}$, $\beta_{\Delta}(d)$ is not uniquely determined, then we choose nondeterministically one possible value.

We can now define the database state $\mathcal{B} = \beta_{\mathcal{S}}(\mathcal{I})$ corresponding to $\mathcal{I}$:

- $\Delta^{\mathcal{B}} = \Delta^{\mathcal{I}} \setminus \left( \bigcup_{R \in \mathcal{R}_{\mathcal{S}}} (\phi(R))^{\mathcal{I}} \cup \bigcup_{D \in \mathcal{D}_{\mathcal{S}}} (\phi(D))^{\mathcal{I}} \right)$.

- For each symbol $X \in \mathcal{E}_{\mathcal{S}} \cup \mathcal{A}_{\mathcal{S}} \cup \mathcal{R}_{\mathcal{S}} \cup \mathcal{D}_{\mathcal{S}}$, $X^{\mathcal{B}} = \beta_{\Delta}((\phi(X))^{\mathcal{I}})$.

It is not difficult to see, that if $\mathcal{I}$ is a model of $\phi(\mathcal{S})$, then $\mathcal{B}$ defined in such a way is a legal database state for $\mathcal{S}$ with active domain $\bigcup_{D \in \mathcal{D}_{\mathcal{S}}} (\phi(D))^{\mathcal{I}}$. $\qquad \square$

The following theorem allows us to reduce reasoning on ER schemata to finite model reasoning on ALUNI knowledge bases.

**Theorem 4.9** *Let $\mathcal{S}$ be an ER schema, $E, E'$ be two entities in $\mathcal{S}$, and $\phi(\mathcal{S})$ be the translation of $\mathcal{S}$. Then the following holds:*

1. *$E$ is satisfiable in $\mathcal{S}$ if and only if $\phi(\mathcal{S}) \not\models_f \phi(E) \preceq \bot$.*

2. *$E$ inherits from $E'$ in $\mathcal{S}$ if and only if $\phi(\mathcal{S}) \models_f \phi(E) \preceq \phi(E')$.*

*Proof.* (1) "⇒" Let $\mathcal{B}$ be a legal database state with $E^{\mathcal{B}} \neq \emptyset$. By part 1 of Proposition 4.8, $\alpha_{\mathcal{S}}(\mathcal{B})$ is a finite model of $\phi(\mathcal{S})$ in which $(\phi(E))^{\alpha_{\mathcal{S}}(\mathcal{B})} \neq \emptyset$.

"⇐" Let $\phi(E)$ be finitely consistent in $\phi(\mathcal{S})$. By Lemma 4.7 there is a finite relation-descriptive model $\mathcal{I}$ of $\phi(\mathcal{S})$ with $\phi(E)^{\mathcal{I}} \neq \emptyset$. By part 2 of Proposition 4.8, $\beta_{\mathcal{S}}(\mathcal{I})$ is a database state legal for $\mathcal{S}$ in which $E^{\mathcal{B}} \neq \emptyset$.

(2) "⇒" Let $\phi(\mathcal{S}) \not\models_f \phi(E) \preceq \phi(E')$. Then $\phi(E) \sqcap \neg\phi(E')$ is finitely consistent in $\phi(\mathcal{S})$. By Lemma 4.7 there is a finite relation-descriptive model $\mathcal{I}$ of $\phi(\mathcal{S})$ with $d \in (\phi(E))^{\mathcal{I}}$ and $d \notin (\phi(E'))^{\mathcal{I}}$, for some $d \in \Delta^{\mathcal{I}}$. By part 2 of Proposition 4.8, $\beta_{\mathcal{S}}(\mathcal{I})$ is a database state legal for $\mathcal{S}$ in which $d \in E^{\mathcal{B}}$ and $d \notin E'^{\mathcal{B}}$. Therefore $E$ does not inherit from $E'$.

"⇐" Assume $E$ does not inherit from $E'$. Then there is a database state $\mathcal{B}$ legal for $\mathcal{S}$ where for an instance $e \in E^{\mathcal{B}}$ we have $e \notin E'^{\mathcal{B}}$. By part 1 of Proposition 4.8, $\alpha_{\mathcal{S}}(\mathcal{B})$ is a finite model of $\phi(\mathcal{S})$ in which $e \in (\phi(E))^{\alpha_{\mathcal{S}}(\mathcal{B})}$ and $e \notin (\phi(E'))^{\alpha_{\mathcal{S}}(\mathcal{B})}$. Therefore $\phi(\mathcal{S}) \not\models_f \phi(E) \preceq \phi(E')$. $\qquad\blacksquare$

Theorem 4.9 allows us to effectively exploit the reasoning methods that have been developed for ALUNI in order to reason on ER schemas. The complexity of the resulting method for reasoning on ER schemata is exponential. Observe however, that the known algorithms for reasoning on ER schemata are also exponential (Calvanese & Lenzerini, 1994b), and that the precise computational complexity of the problem is still open.

Moreover, by exploiting the correspondence with ALUNI, it becomes possible to add to the ER model (and more in general to semantic data models) several features and modeling primitives that are currently missing, and which have been considered important, and fully take them into account when reasoning over schemata. Such additional features include for example the possibility to specify and use arbitrary boolean combinations of entities, and to refine properties of entities along ISA hierarchies.

## 5. Object-Oriented Data Models

Object-oriented data models have been proposed with the goal of devising database formalisms that could be integrated with object-oriented programming systems (Kim, 1990). They are the subject of an active area of research in the database field, and are based on the following features:

- They rely on the notion of object identifier at the extensional level (as opposed to traditional data models which are value-oriented) and on the notion of class at the intensional level.

- The structure of the classes is specified by means of typing and inheritance.

As in the previous section, we present the common basis of object-oriented data models with other class-based formalisms by introducing a language for specifying object-oriented schemata and show that such schemata can be correctly represented as ALUNI knowledge bases. In our analysis, we concentrate our attention on the structural aspects of object-oriented data models. One of the characteristics of the object-oriented approach is to provide mechanisms for specifying also the dynamic properties of classes and objects, typically through the definition of methods associated to the classes. Those aspects are outside the scope of our investigations. Nevertheless, we argue that general techniques for schema level reasoning, in particular, type consistency and type inference, can be profitably exploited for restricted forms of reasoning on methods (Abiteboul, Kanellakis, Ramaswamy, & Waller, 1992).

## 5.1 Syntax of an Object-Oriented Model

Below we define a simple object-oriented language in the style of most popular models featuring complex objects and object identity. Although we do not refer to any specific formalism, our model is inspired by the ones presented by Abiteboul and Kanellakis (1989), Hull and King (1987).

**Definition 5.1** An *object-oriented schema* is a tuple $\mathcal{S} = (\mathcal{C}_\mathcal{S}, \mathcal{A}_\mathcal{S}, \mathcal{D}_\mathcal{S})$, where:

- $\mathcal{C}_\mathcal{S}$ is a finite set of *class names*, denoted by the letter $C$.
- $\mathcal{A}_\mathcal{S}$ is a finite set of *attribute names*, denoted by the letter $A$.
- $\mathcal{D}_\mathcal{S}$ is a finite set of *class declarations* of the form

$$\underline{\text{Class}}\ C\ \underline{\text{is-a}}\ C_1, \ldots, C_k\ \underline{\text{type-is}}\ T,$$

in which $T$ denotes a *type expression* built according to the following syntax:

$$
\begin{aligned}
T \quad \longrightarrow \quad & C\ | \\
& \underline{\text{Union}}\ T_1, \ldots, T_k\ \underline{\text{End}}\ | \\
& \underline{\text{Set-of}}\ T\ | \\
& \underline{\text{Record}}\ A_1{:}T_1, \ldots, A_k{:}T_k\ \underline{\text{End}}.
\end{aligned}
$$

$\mathcal{D}_\mathcal{S}$ contains exactly one such declaration for each class $C \in \mathcal{C}_\mathcal{S}$.  ■

**Example 5.2** Figure 7 shows a fragment of the object-oriented schema corresponding to the KEE knowledge base of Figure 2.  ■

Each class declaration imposes constraints on the instances of the class it refers to. The is-a part of a class declaration allows one to specify inclusion between the sets of instances of the involved classes, while the type-is part specifies through a type expression the structure assigned to the objects that are instances of the class.

```
Class Teacher type-is          Class Course type-is
   Union Professor, GradStudent    Record
   End                                enrolls: Set-of Student,
                                      taughtby: Teacher
Class GradStudent is-a Student type-is  End
   Record
      degree: String
   End
```

Figure 7: An object-oriented schema

## 5.2 Semantics of an Object-Oriented Model

The meaning of an object-oriented schema is given by specifying the characteristics of an instance of the schema. The definition of instance makes use of the notions of object identifier and value.

Let us first characterize the set of values that can be constructed from a set of symbols, called *object identifiers*. Given a finite set $\mathcal{O}$ of symbols denoting real world objects, the set $\mathcal{V}_{\mathcal{O}}$ of *values* over $\mathcal{O}$ is inductively defined as follows:

- $\mathcal{O} \subseteq \mathcal{V}_{\mathcal{O}}$.
- If $v_1, \ldots, v_k \in \mathcal{V}_{\mathcal{O}}$ then $\{| v_1, \ldots, v_k |\} \in \mathcal{V}_{\mathcal{O}}$.
- If $v_1, \ldots, v_k \in \mathcal{V}_{\mathcal{O}}$ then $[\![ A_1 : v_1, \ldots, A_k : v_k ]\!] \in \mathcal{V}_{\mathcal{O}}$.
- Nothing else is in $\mathcal{V}_{\mathcal{O}}$.

A *database instance* $\mathcal{J}$ of a schema $\mathcal{S} = (\mathcal{C}_{\mathcal{S}}, \mathcal{A}_{\mathcal{S}}, \mathcal{D}_{\mathcal{S}})$ is constituted by

- a *finite* set $\mathcal{O}^{\mathcal{J}}$ of object identifiers;
- a mapping $\pi^{\mathcal{J}}$ assigning to each class in $\mathcal{C}_{\mathcal{S}}$ a subset of $\mathcal{O}^{\mathcal{J}}$;
- a mapping $\rho^{\mathcal{J}}$ assigning a value in $\mathcal{V}_{\mathcal{O}^{\mathcal{J}}}$ to each object in $\mathcal{O}^{\mathcal{J}}$.

Although the set $\mathcal{V}_{\mathcal{O}^{\mathcal{J}}}$ of values that can be constructed from a set $\mathcal{O}^{\mathcal{J}}$ of object identifiers is infinite, for a database instance one needs only to consider a finite subset of $\mathcal{V}_{\mathcal{O}^{\mathcal{J}}}$.

**Definition 5.3** Given an object-oriented schema $\mathcal{S}$ and an instance $\mathcal{J}$ of $\mathcal{S}$, the set $\mathcal{V}_{\mathcal{J}}$ of *active values* with respect to $\mathcal{J}$ is constituted by:

- the set $\mathcal{O}^{\mathcal{J}}$ of object identifiers.
- the set of values assigned by $\rho^{\mathcal{J}}$ to the elements of $\mathcal{O}^{\mathcal{J}}$, including those values that are not explicitly associated with object identifiers, but are used to form other values. ∎

The interpretation of type expressions in $\mathcal{J}$ is defined through an *interpretation function* $\cdot^{\mathcal{J}}$ that assigns to each type expression a subset of $\mathcal{V}_{\mathcal{O}^{\mathcal{J}}}$ such that the following conditions are satisfied:

$$C^{\mathcal{J}} \;=\; \pi^{\mathcal{J}}(C)$$

$$
\begin{aligned}
(\underline{\text{Union }} T_1, \ldots, T_k \ \underline{\text{End}})^{\mathcal{J}} &= T_1^{\mathcal{J}} \cup \cdots \cup T_k^{\mathcal{J}} \\
(\underline{\text{Set-of }} T)^{\mathcal{J}} &= \{\!\{v_1, \ldots, v_k\}\!\} \mid k \geq 0, v_i \in T^{\mathcal{J}}, \text{ for } i \in \{1, \ldots, k\}\} \\
(\underline{\text{Record }} A_1{:}T_1, \ldots, A_k{:}T_k \ \underline{\text{End}})^{\mathcal{J}} &= \{[\![A_1{:}v_1, \ldots, A_h{:}v_h]\!] \mid h \geq k, \\
& \qquad v_i \in T_i^{\mathcal{J}}, \text{ for } i \in \{1, \ldots, k\}, \\
& \qquad v_j \in \mathcal{V}_{\mathcal{O}^{\mathcal{J}}}, \text{ for } j \in \{k+1, \ldots, h\}\}.
\end{aligned}
$$

Notice that the instances of type record may have more components than those specified in the type of the class. Thus we are using an open semantics for records, which is typical of object-oriented data models (Abiteboul & Kanellakis, 1989).

In order to characterize object-oriented data models we consider the instances that are admissible for the schema.

**Definition 5.4** Let $\mathcal{S} = (\mathcal{C}_{\mathcal{S}}, \mathcal{A}_{\mathcal{S}}, \mathcal{D}_{\mathcal{S}})$ be an object-oriented schema. A database instance $\mathcal{J}$ of $\mathcal{S}$ is said to be *legal* (with respect to $\mathcal{S}$) if for each declaration

$$\underline{\text{Class }} C \ \underline{\text{is-a}} \ C_1, \ldots, C_n \ \underline{\text{type-is}} \ T$$

in $\mathcal{D}_{\mathcal{S}}$, it holds that $C^{\mathcal{J}} \subseteq C_i^{\mathcal{J}}$ for each $i \in \{1, \ldots, n\}$, and that $\rho^{\mathcal{J}}(C^{\mathcal{J}}) \subseteq T^{\mathcal{J}}$. ∎

Therefore, for a legal database instance, the type expressions that are present in the schema determine the (finite) set of active values that must be considered. The construction of such values is limited by the depth of type expressions.

### 5.3 Relationship between Object-Oriented Schemata and ALUNI

We establish now a relationship between ALUNI and the object-oriented language presented above. This is done by providing a mapping from object-oriented schemata into ALUNI knowledge bases. Since the interpretation domain for ALUNI knowledge bases consists of atomic objects, whereas each instance of an object-oriented schema is assigned a possibly structured value (see the definition of $\mathcal{V}_{\mathcal{O}}$), we need to explicitly represent some of the notions that underlie the object-oriented language. In particular, while there is a correspondence between concepts and classes, one must explicitly account for the type structure of each class. This can be accomplished by introducing in ALUNI concepts `AbstractClass`, to represent the classes, and `RecType` and `SetType` to represent the corresponding types. The associations between classes and types induced by the class declarations, as well as the basic characteristics of types, are modeled by means of roles: the (functional) role `value` models the association between classes and types, and the role `member` is used for specifying the type of the elements of a set. Moreover, the concepts representing types are assumed to be mutually disjoint, and disjoint from the concepts representing classes. These constraints are expressed by adequate inclusion assertions that will be part of the knowledge base we are going to define.

We first define the function $\psi$ that maps each type expression into an $\mathcal{ALUNI}$ concept expression as follows:

- Every class $C$ is mapped into an atomic concept $\psi(C)$.

- Every type expression $\underline{\text{Union }} T_1, \ldots, T_k \ \underline{\text{End}}$ is mapped into $\psi(T_1) \sqcup \cdots \sqcup \psi(T_k)$.

- Every type expression <u>Set-of</u> $T$ is mapped into $\texttt{SetType} \sqcap \forall \texttt{member}.\psi(T)$.

- Every attribute $A$ is mapped into an atomic role $\psi(A)$, and every type expression <u>Record</u> $A_1 : T_1, \ldots, A_k : T_k$ <u>End</u> is mapped into

$$\texttt{RecType} \sqcap \forall\psi(A_1).\psi(T_1) \sqcap \exists^{=1}\psi(A_1) \sqcap \cdots \sqcap$$
$$\forall\psi(A_k).\psi(T_k) \sqcap \exists^{=1}\psi(A_k).$$

Using $\psi$ we define the ALUNI knowledge base corresponding to an object-oriented schema.

**Definition 5.5** The ALUNI knowledge base $\psi(\mathcal{S}) = (\mathcal{A}, \mathcal{P}, \mathcal{T})$ corresponding to the object-oriented schema $\mathcal{S} = (\mathcal{C}_\mathcal{S}, \mathcal{A}_\mathcal{S}, \mathcal{D}_\mathcal{S})$ is obtained as follows:

- $\mathcal{A} = \{\texttt{AbstractClass}, \texttt{RecType}, \texttt{SetType}\} \cup \{\psi(C) \mid C \in \mathcal{C}_\mathcal{S}\}$.

- $\mathcal{P} = \{\texttt{value}, \texttt{member}\} \cup \{\psi(A) \mid A \in \mathcal{A}_\mathcal{S}\}$.

- $\mathcal{T}$ consists of the following assertions:

$$\begin{aligned}
\texttt{AbstractClass} &\;\dot{\preceq}\; \exists^{=1}\texttt{value} \\
\texttt{RecType} &\;\dot{\preceq}\; \forall\texttt{value}.\bot \\
\texttt{SetType} &\;\dot{\preceq}\; \forall\texttt{value}.\bot \sqcap \neg\texttt{RecType}
\end{aligned}$$

  and for each class declaration

$$\underline{\text{Class}}\ C\ \underline{\text{is-a}}\ C_1, \ldots, C_n\ \underline{\text{type-is}}\ T$$

  in $\mathcal{D}_\mathcal{S}$, an inclusion assertion

$$\psi(C) \;\dot{\preceq}\; \texttt{AbstractClass} \sqcap \psi(C_1) \sqcap \cdots \sqcap \psi(C_n) \sqcap \forall\texttt{value}.\psi(T).$$

∎

From the above translation we can observe that inverse roles are not necessary for the formalization of object-oriented data models. Indeed, the possibility of referring to the inverse of an attribute is generally ruled out in such models. However, this strongly limits the expressive power of the data model, as pointed out in recent papers (see for example Albano, Ghelli, & Orsini, 1991; Cattell, 1994). Note also that the use of number restrictions is limited to the value 1, which corresponds to existence constraints and functionality, whereas union is used in a more general form than for example in the KEE system.

**Example 5.2 (cont.)** We illustrate the translation on the fragment of object-oriented schema in Figure 7. The corresponding ALUNI knowledge base is shown in Figure 8. ∎

$\mathcal{K} = (\mathcal{A}, \mathcal{P}, \mathcal{T})$, where

$\mathcal{A} = \{\texttt{AbstractClass}, \texttt{RecType}, \texttt{SetType}, \texttt{String},$
$\qquad \texttt{Course}, \texttt{Teacher}, \texttt{Professor}, \texttt{Student}, \texttt{GradStudent}\},$

$\mathcal{P} = \{\texttt{value}, \texttt{member}, \texttt{enrolls}, \texttt{taughtby}, \texttt{degree}\},$

and the set $\mathcal{T}$ of assertions consists of:

$$
\begin{aligned}
\texttt{Course} \;\dot{\preceq}\;& \texttt{AbstractClass} \sqcap \\
& \forall\texttt{value}.(\texttt{RecType} \sqcap \exists^{=1}\texttt{enrolls} \sqcap \exists^{=1}\texttt{taughtby} \sqcap \\
& \qquad \forall\texttt{enrolls}.(\texttt{SetType} \sqcap \forall\texttt{member}.\texttt{Student}) \sqcap \forall\texttt{taughtby}.\texttt{Teacher}) \\
\texttt{Teacher} \;\dot{\preceq}\;& \texttt{AbstractClass} \sqcap \forall\texttt{value}.(\texttt{GradStudent} \sqcup \texttt{Professor}) \\
\texttt{GradStudent} \;\dot{\preceq}\;& \texttt{AbstractClass} \sqcap \texttt{Student} \sqcap \\
& \forall\texttt{value}.(\texttt{RecType} \sqcap \forall\texttt{degree}.\texttt{String} \sqcap \exists^{=1}\texttt{degree}) \\
\texttt{AbstractClass} \;\dot{\preceq}\;& \exists^{=1}\texttt{value} \\
\texttt{RecType} \;\dot{\preceq}\;& \forall\texttt{value}.\bot \\
\texttt{SetType} \;\dot{\preceq}\;& \forall\texttt{value}.\bot \sqcap \neg\texttt{RecType}
\end{aligned}
$$

Figure 8: The ALUNI knowledge base corresponding to the object-oriented schema in Figure 7

Below we discuss the effectiveness of the translation $\psi$. First of all observe that the ALUNI knowledge base $\psi(\mathcal{S})$ resulting from the translation of an object-oriented schema $\mathcal{S}$ may admit models that do not have a direct counterpart among legal database instances of $\mathcal{S}$. More precisely, both an interpretation of $\psi(\mathcal{S})$ and a database instance of $\mathcal{S}$ can be viewed as a directed labeled graph: In the case of an interpretation, the nodes are domain individuals and the arcs are labeled with roles. In the case of a database instance, the nodes are either object identifiers or active values, and an arc either connects an object identifier to its associated value (in which case it is labeled with value), or is part of the sub-structure representing a set or record value (in which case it is labeled with member or with an attribute, in accordance with the type of the value). In a legal database instance of $\mathcal{S}$, a value $v$ is represented by a sub-structure that has the form of a finite tree with $v$ as root, set and record values as intermediate nodes, and objects identifiers as leaves. Clearly, such a substructure does not contain cycles. Conversely, in a model of $\psi(\mathcal{S})$, there may be cycles involving only nodes that are instances of SetType and RecType and in which all roles are different from value. We call such cycles *bad*. A model containing bad cycles cannot be put directly in correspondence with a legal database instance. Also, due to the open semantics of records one cannot adopt a different translation for which bad cycles in the model are ruled out.

**Example 5.6** Consider the object-oriented schema $\mathcal{S}$, containing a single class declaration

<u>Class</u> $C$ <u>type-is</u> <u>Record</u> $a_1$ : <u>Record</u> $a_2$ : <u>Record</u> $a_3$ : $C$ <u>End</u> <u>End</u> <u>End</u>

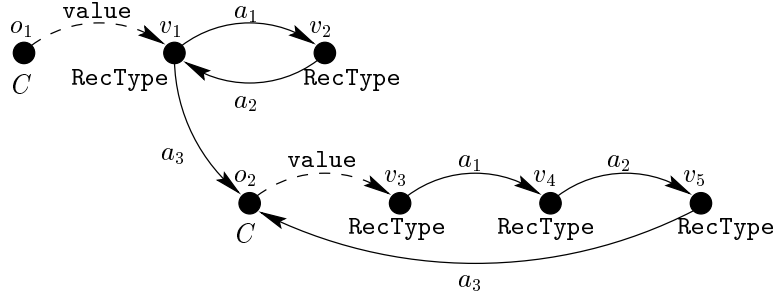

Figure 9: A model containing cycles

which is translated to

$$C \ \dot{\preceq} \ \texttt{AbstractClass} \sqcap$$
$$\forall \texttt{value}.(\texttt{RecType} \sqcap \exists^{=1}a_1 \sqcap \forall a_1.(\texttt{RecType} \sqcap \exists^{=1}a_2 \sqcap \forall a_2.(\texttt{RecType} \sqcap \exists^{=1}a_3 \sqcap \forall a_3.C))).$$

Figure 9 shows a model of $\psi(\mathcal{S})$ represented as a graph. For clarity, we have named the instances of $C$, and hence of $\texttt{AbstractClass}$, with $o$ and the instances of $\texttt{RecType}$ with $v$. Observe the two different types of cycles in the graph. The cycle involving individuals $o_2, v_3, v_4$, and $v_5$ does not cause any problems since it contains an arc labeled with $\texttt{value}$, which is not part of the structure constituting a complex value. In fact, $v_3$ represents the record value $[\![a_1\!:\![\![a_2\!:\![\![a_3\!:o_2]\!]]\!]]\!]$. On the other hand, due to the bad cycle involving $v_1$ and $v_2$, individual $v1$ represents (together with $o_2$ connected via $a_3$ to $v_1$) a record of infinite depth. ∎

We can nevertheless establish a correspondence from finite models of $\psi(\mathcal{S})$ possibly containing bad cycles to legal instances of the object-oriented schema $\mathcal{S}$. This can be achieved by unfolding the bad cycles in a model of $\psi(\mathcal{S})$ to infinite trees. Obviously, the unfolding of a cycle into an infinite tree, generates an infinite number of nodes, which would correspond to an infinite database state. However, we can restrict the duplication of individuals to those that represent set and record values, and thus are instances of $\texttt{SetType}$ and $\texttt{RecType}$. The instances of $\texttt{AbstractClass}$, instead, are not duplicated in the process of unfolding, and therefore their number remains finite. Moreover, since the set of possible active values associated with each object identifier is bound by the depth of the schema, we can in fact block the unfolding of bad cycles to the finite tree of depth equal to the depth of the schema.

Let us first formally define the *depth* of an object-oriented schema $\mathcal{S}$.

**Definition 5.7** For a type expression $T$ we define $depth(T)$ inductively as follows:

$$depth(T) = \begin{cases} 0, & \text{if } T = C. \\ \max_{1 \le i \le k}(depth(T_i)), & \text{if } T = \underline{\text{Union}}\ T_1, \ldots, T_k\ \underline{\text{End}}. \\ 1 + depth(T'), & \text{if } T = \underline{\text{Set-of}}\ T'. \\ 1 + \max_{1 \le i \le k}(depth(T_i)), & \text{if } T = \underline{\text{Record}}\ A_1\!:T_1, \ldots, A_k\!:T_k\ \underline{\text{End}}. \end{cases}$$

The *depth* of an object-oriented schema $\mathcal{S}$ is defined as the maximum of $depth(T)$ for a type expression $T$ in $\mathcal{S}$. ∎

227

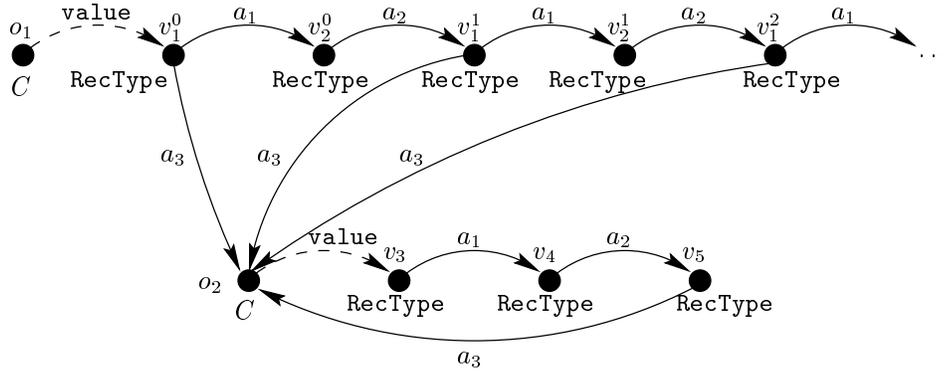

Figure 10: The unfolded version of the model in Figure 9

We can now introduce the notion of unfolding of an ALUNI interpretation.

**Definition 5.8** Let $\mathcal{S}$ be an object-oriented schema, $\psi(\mathcal{S})$ its translation in ALUNI and $\mathcal{I}$ a finite interpretation of $\psi(\mathcal{S})$. We call *unfolded version* of $\mathcal{I}$ the interpretation obtained from $\mathcal{I}$ as follows: For each individual $v$ that is part of a bad cycle, unfold the bad cycle into an (infinite) tree having $v$ as root, by generating new individuals only for the instances of RecType and SetType. For a nonnegative integer $m$, we call *m-unfolded version* of $\mathcal{I}$, denoted as $\mathcal{I}_{|m}$, the interpretation obtained by truncating at depth $m$ each infinite tree generated in the process of unfolding. ∎

**Example 5.6 (cont.)** Figure 10 shows the unfolded version of the model in Figure 9. Notice that only the bad cycle has been unfolded to an infinite tree, and that all arcs labeled with $a_3$ lead to $o_2$, which is an instance of AbstractClass and has not been duplicated. ∎

The correctness of $\psi(\mathcal{S})$ is sanctioned by the following proposition.

**Proposition 5.9** *For every object-oriented schema $\mathcal{S}$ of depth $m$, there exist mappings:*

1. *$\alpha_{\mathcal{S}}$ from instances of $\mathcal{S}$ into finite interpretations of $\psi(\mathcal{S})$ and $\alpha_{\mathcal{V}}$ from active values of instances of $\mathcal{S}$ into domain elements of the finite interpretations of $\psi(\mathcal{S})$ such that: For each legal instance $\mathcal{J}$ of $\mathcal{S}$, $\alpha_{\mathcal{S}}(\mathcal{J})$ is a finite model of $\psi(\mathcal{S})$, and for each type expression $T$ of $\mathcal{S}$ and each $v \in \mathcal{V}_{\mathcal{J}}$, $v \in T^{\mathcal{J}}$ if and only if $\alpha_{\mathcal{V}}(v) \in (\psi(T))^{\alpha_{\mathcal{S}}(\mathcal{J})}$.*

2. *$\beta_{\mathcal{S}}$ from finite interpretations of $\psi(\mathcal{S})$ into instances of $\mathcal{S}$ and $\beta_{\mathcal{V}}$ from domain elements of the $m$-unfolded versions of the finite interpretations of $\psi(\mathcal{S})$ into active values of instances of $\mathcal{S}$, such that: For each finite model $\mathcal{I}$ of $\psi(\mathcal{S})$, $\beta_{\mathcal{S}}(\mathcal{I})$ is a legal instance of $\mathcal{S}$, and for each concept $\psi(T)$, which is the translation of a type expression $T$ of $\mathcal{S}$ and each $d \in \Delta^{\mathcal{I}_{|m}}$, $d \in (\psi(T))^{\mathcal{I}_{|m}}$ if and only if $\beta_{\mathcal{V}}(d) \in T^{\beta_{\mathcal{S}}(\mathcal{I})}$.*

*Proof.* (1) Given a database instance $\mathcal{J}$ we define an interpretation $\alpha_{\mathcal{S}}(\mathcal{J})$ of $\psi(\mathcal{S})$ as follows:

- $\alpha_{\mathcal{V}}$ is a function mapping every element of $\mathcal{V}_{\mathcal{J}}$ into a distinct element of $\Delta^{\alpha_{\mathcal{S}}(\mathcal{J})}$. Therefore $\Delta^{\alpha_{\mathcal{S}}(\mathcal{J})}$ is defined as the set of elements $\alpha_{\mathcal{V}}(v)$ such that $v \in \mathcal{V}_{\mathcal{J}}$. Moreover we denote with $\Delta_{id}$, $\Delta_{rec}$, and $\Delta_{set}$ the elements of $\Delta^{\alpha_{\mathcal{S}}(\mathcal{J})}$ corresponding to object identifiers, record and set values, respectively.

- The interpretation of atomic concepts is defined as follows:

$$
\begin{aligned}
(\psi(C))^{\alpha_{\mathcal{S}}(\mathcal{J})} &= \{\alpha_{\mathcal{V}}(o) \mid o \in \pi^{\mathcal{J}}(C)\}, \\
&\quad \text{for every } \psi(C) \text{ corresponding to a class name } C \text{ in } \mathcal{S} \\
\texttt{AbstractClass}^{\alpha_{\mathcal{S}}(\mathcal{J})} &= \Delta_{id} \\
\texttt{RecType}^{\alpha_{\mathcal{S}}(\mathcal{J})} &= \Delta_{rec} \\
\texttt{SetType}^{\alpha_{\mathcal{S}}(\mathcal{J})} &= \Delta_{set}
\end{aligned}
$$

- The interpretation of atomic roles is defined as follows:

$$
\begin{aligned}
(\psi(A))^{\alpha_{\mathcal{S}}(\mathcal{J})} &= \{(d_1, d_2) \mid d_1 \in \Delta_{rec} \text{ and } \alpha_{\mathcal{V}}^{-1}(d_1) = [\![\ldots, A{:}\alpha_{\mathcal{V}}^{-1}(d_2), \ldots]\!]\}, \\
&\quad \text{for every } \psi(A) \text{ corresponding to an attribute name } A \text{ in } \mathcal{S} \\
\texttt{member}^{\alpha_{\mathcal{S}}(\mathcal{J})} &= \{(d_1, d_2) \mid d_1 \in \Delta_{set} \text{ and } \alpha_{\mathcal{V}}^{-1}(d_1) = \{\!|\ldots, \alpha_{\mathcal{V}}^{-1}(d_2), \ldots|\!\}\} \\
\texttt{value}^{\alpha_{\mathcal{S}}(\mathcal{J})} &= \{(d_1, d_2) \mid (\alpha_{\mathcal{V}}^{-1}(d_1), \alpha_{\mathcal{V}}^{-1}(d_2)) \in \rho^{\mathcal{J}}\}
\end{aligned}
$$

We prove that for each type $T$ and each $v \in \mathcal{V}_{\mathcal{J}}$, $v \in T^{\mathcal{J}}$ if and only if $\alpha_{\mathcal{V}}(v) \in (\psi(T))^{\alpha_{\mathcal{S}}(\mathcal{J})}$. The first part of the thesis then follows from the definition of $\alpha_{\mathcal{S}}(\mathcal{J})$. The proof is by induction on the structure of the type expression.

Base case: $T = C$ (i.e., $T$ is a class name). If $o \in C^{\mathcal{J}}$ then $\alpha_{\mathcal{V}}(o) \in (\psi(C))^{\alpha_{\mathcal{S}}(\mathcal{J})}$, and vice-versa if $d \in (\psi(C))^{\alpha_{\mathcal{S}}(\mathcal{J})}$ then $\alpha_{\mathcal{V}}^{-1}(d) \in C^{\mathcal{J}}$.

Inductive case: $T = \underline{\text{Record}}\ A_1{:}T_1, \ldots, A_k{:}T_k\ \underline{\text{End}}$ and $\psi(T) = \texttt{RecType} \sqcap \forall \psi(A_1).\psi(T_1) \sqcap \cdots \sqcap \forall \psi(A_k).\psi(T_k) \sqcap \exists^{=1}\psi(A_1) \sqcap \cdots \sqcap \exists^{=1}\psi(A_k)$. We assume that $v \in T_i^{\mathcal{J}}$ iff $\alpha_{\mathcal{V}}(v) \in (\psi(T_i))^{\alpha_{\mathcal{S}}(\mathcal{J})}$, for $i \in \{1, \ldots, k\}$, and show that $v \in T^{\mathcal{J}}$ iff $\alpha_{\mathcal{V}}(v) \in (\psi(T))^{\alpha_{\mathcal{S}}(\mathcal{J})}$.

Suppose that $v \in T^{\mathcal{J}}$, i.e., $v = [\![A_1{:}v_1, \ldots, A_h{:}v_h]\!]$ with $h \geq k$ and $v_i \in T_i^{\mathcal{J}}$ for $i \in \{1, \ldots, k\}$. By induction hypothesis $\alpha_{\mathcal{V}}(v_i) \in (\psi(T_i))^{\alpha_{\mathcal{S}}(\mathcal{J})}$, for $i \in \{1, \ldots, k\}$, and by definition of $\alpha_{\mathcal{S}}$, $\alpha_{\mathcal{V}}(v) \in \texttt{RecType}^{\alpha_{\mathcal{S}}(\mathcal{J})}$, $(\alpha_{\mathcal{V}}(v), \alpha_{\mathcal{V}}(v_i)) \in (\psi(A_i))^{\alpha_{\mathcal{S}}(\mathcal{J})}$ for $i \in \{1, \ldots, k\}$, and all roles $\psi(A)$ corresponding to attribute names are functional. Therefore, $\alpha_{\mathcal{V}}(v) \in (\psi(T))^{\alpha_{\mathcal{S}}(\mathcal{J})}$.

Conversely, suppose that $d = \alpha_{\mathcal{V}}(v) \in (\psi(T))^{\alpha_{\mathcal{S}}(\mathcal{J})}$. Then, for each $i \in \{1, \ldots, k\}$ there is exactly one $d_i \in \Delta^{\alpha_{\mathcal{S}}(\mathcal{J})}$ such that $(d, d_i) \in (\psi(A_i))^{\alpha_{\mathcal{S}}(\mathcal{J})}$, and moreover $d_i \in (\psi(T_i))^{\alpha_{\mathcal{S}}(\mathcal{J})}$. By definition of $\alpha_{\mathcal{S}}$ we have $v = [\![A_1{:}v_1, \ldots, A_h{:}v_h]\!]$, with $h \geq k$ and $v_i = \alpha_{\mathcal{V}}^{-1}(d_i)$, for $i \in \{1, \ldots, k\}$. By induction hypothesis $v_i \in T_i^{\mathcal{J}}$, for $i \in \{1, \ldots, k\}$, and therefore $v \in (\underline{\text{Record}}\ A_1{:}T_1, \ldots, A_k{:}T_k\ \underline{\text{End}})^{\mathcal{J}}$.

The cases for $T = \underline{\text{Union}}\ T_1, \ldots, T_k\ \underline{\text{End}}$ and $T = \underline{\text{Set-of}}\ T'$ can be treated analogously.

(2) Given a finite model $\mathcal{I}$ of $\psi(\mathcal{S})$ of depth $m$, we define a legal database instance $\beta_{\mathcal{S}}(\mathcal{I})$ as follows:

- $\beta_{\mathcal{V}}$ is a function mapping every element of $\Delta^{\mathcal{I}|_m}$ into a distinct element of $\mathcal{V}_{\beta_{\mathcal{S}}(\mathcal{I})}$ such that the following conditions are satisfied:

  - $\mathcal{O}^{\beta_{\mathcal{S}}(\mathcal{I})} \subseteq \mathcal{V}_{\beta_{\mathcal{S}}(\mathcal{I})}$ is the set of elements $\beta_{\mathcal{V}}(d)$ such that $d \in \texttt{AbstractClass}^{\mathcal{I}|_m}$.

- If $d \in \mathtt{RecType}^{\mathcal{I}_{|m}}$, $(d, d_i) \in (\psi(A_i))^{\mathcal{I}_{|m}}$, for $i \in \{1, \ldots, k\}$, and there is no other individual $d' \in \Delta^{\mathcal{I}_{|m}}$ and attribute $A'$ such that $(d, d') \in (\psi(A'))^{\mathcal{I}_{|m}}$, then $\beta_{\mathcal{V}}(d) = [\![A_1 \colon \beta_{\mathcal{V}}(d_1), \ldots, A_k \colon \beta_{\mathcal{V}}(d_k)]\!]$.

- If $d \in \mathtt{SetType}^{\mathcal{I}_{|m}}$, $(d, d_i) \in \mathtt{member}^{\mathcal{I}_{|m}}$, for $i \in \{1, \ldots, k\}$, and there is no other individual $d' \in \Delta^{\mathcal{I}_{|m}}$ such that $(d, d') \in (\mathtt{member})^{\mathcal{I}_{|m}}$, then $\beta_{\mathcal{V}}(d) = \{\beta_{\mathcal{V}}(d_1), \ldots, \beta_{\mathcal{V}}(d_k)\}$.

- For every class name $C$, $\pi^{\beta_{\mathcal{S}}(\mathcal{I})}(C) = \{\beta_{\mathcal{V}}(d) \mid d \in (\psi(C))^{\mathcal{I}_{|m}}\}$.

- $\rho^{\beta_{\mathcal{S}}(\mathcal{I})} = \{(o, v) \mid \beta_{\mathcal{V}}(d_1) = o, \beta_{\mathcal{V}}(d_2) = v, \text{ and } (d_1, d_2) \in \mathtt{value}^{\mathcal{I}_{|m}}\}$.

We first prove that for each concept $\psi(T)$, which is the translation of a type expression $T$ of $\mathcal{S}$, and each $d \in \Delta^{\mathcal{I}_{|m}}$, $d \in (\psi(T))^{\mathcal{I}_{|m}}$ if and only if $\beta_{\mathcal{V}}(d) \in T^{\beta_{\mathcal{S}}(\mathcal{I})}$. The proof is by induction on the structure of the concept expression. Again for the inductive part we restrict our attention to the case of record types.

Base case: $T = C$ (i.e., $\psi(T)$ is an atomic concept). If $d \in (\psi(C))^{\mathcal{I}_{|m}}$ then $\beta_{\mathcal{V}}(d) \in C^{\beta_{\mathcal{S}}(\mathcal{I})}$, and vice-versa if $o \in C^{\beta_{\mathcal{S}}(\mathcal{I})}$ then $\beta_{\mathcal{V}}^{-1}(o) \in (\psi(C))^{\mathcal{I}_{|m}}$.

Inductive case: $\psi(T) = \mathtt{RecType} \sqcap \forall \psi(A_1).\psi(T_1) \sqcap \exists^{=1} \psi(A_1) \sqcap \cdots \sqcap \forall \psi(A_k).\psi(T_k) \sqcap \exists^{=1} \psi(A_k)$ and $T = \underline{\mathrm{Record}}\ A_1 \colon T_1, \ldots, A_k \colon T_k\ \underline{\mathrm{End}}$. We assume that $d \in (\psi(T_i))^{\mathcal{I}_{|m}}$ iff $\beta_{\mathcal{V}}(d) \in T_i^{\beta_{\mathcal{S}}(\mathcal{I})}$, for $i \in \{1, \ldots, k\}$, and show that $d \in (\psi(T))^{\mathcal{I}_{|m}}$ iff $\beta_{\mathcal{V}}(d) \in T^{\beta_{\mathcal{S}}(\mathcal{I})}$.

Suppose that $d \in (\psi(T))^{\mathcal{I}_{|m}}$. Then $d \in \mathtt{RecType}^{\mathcal{I}_{|m}}$ and for each $i \in \{1, \ldots, k\}$ there is an individual $d_i$ such that $d_i \in (\psi(T_i))^{\mathcal{I}_{|m}}$ and $(d, d_i) \in (\psi(A_i))^{\mathcal{I}_{|m}}$. By construction $\beta_{\mathcal{V}}(d) = [\![A_1 \colon v_1, \ldots, A_h \colon v_h]\!]$ for some $h \geq k$. Moreover, by induction hypothesis $\beta_{\mathcal{V}}(d_i) \in T_i^{\beta_{\mathcal{S}}(\mathcal{I})}$ and therefore $\beta_{\mathcal{V}}(d) \in T^{\beta_{\mathcal{S}}(\mathcal{I})}$.

Conversely, suppose that $\beta_{\mathcal{V}}(d) \in T^{\beta_{\mathcal{S}}(\mathcal{I})}$, i.e., $\beta_{\mathcal{V}}(d) = [\![A_1 \colon v_1, \ldots, A_h \colon v_h]\!]$ with $h \geq k$ and $v_i \in T_i^{\beta_{\mathcal{S}}(\mathcal{I})}$ for $i \in \{1, \ldots, k\}$. By induction hypothesis $d_i = \beta_{\mathcal{V}}^{-1}(v_i) \in (\psi(T_i))^{\mathcal{I}_{|m}}$, for $i \in \{1, \ldots, k\}$, and by definition of $\beta_{\mathcal{V}}$, $d \in \mathtt{RecType}^{\mathcal{I}_{|m}}$ and $(d, d_i) \in (\psi(A_i))^{\mathcal{I}_{|m}}$, for $i \in \{1, \ldots, k\}$. Since all roles $\psi(A)$ corresponding to attribute names are functional, $d \in (\psi(T))^{\mathcal{I}_{|m}}$.

It remains to show that for each declaration

$$\underline{\mathrm{Class}}\ C\ \underline{\mathrm{is\text{-}a}}\ C_1, \ldots, C_n\ \underline{\mathrm{type\text{-}is}}\ T$$

in $\mathcal{D}_{\mathcal{S}}$, (a) $C^{\beta_{\mathcal{S}}(\mathcal{I})} \subseteq C_i^{\beta_{\mathcal{S}}(\mathcal{I})}$ for each $i \in \{1, \ldots, n\}$, and (b) $\rho^{\beta_{\mathcal{S}}(\mathcal{I})}(C^{\beta_{\mathcal{S}}(\mathcal{I})}) \subseteq T^{\beta_{\mathcal{S}}(\mathcal{I})}$.

(a) follows from the fact that $\psi(\mathcal{S})$ contains the assertion $\psi(C) \stackrel{.}{\preceq} \psi(C_1) \sqcap \cdots \sqcap \psi(C_n)$ and from the definition of $\pi^{\beta_{\mathcal{S}}(\mathcal{I})}$.

(b) follows from what we have shown above and from the fact that $\mathcal{I}_{|m}$ still satisfies the assertion $\psi(C) \stackrel{.}{\preceq} \mathtt{AbstractClass} \sqcap \forall \mathtt{value}.\psi(T)$. In fact, for some $d \in (\psi(C))^{\mathcal{I}}$ let $d'$ be the unique individual such that $(d, d') \in \mathtt{value}^{\mathcal{I}}$. Since $\mathcal{I}$ is a model of $\psi(\mathcal{S})$, $d' \in (\psi(T))^{\mathcal{I}}$. We argue that also $d' \in (\psi(T))^{\mathcal{I}_{|m}}$. If $d'$ is not part of a bad cycle in $\mathcal{I}$, then $\mathcal{I}$ and $\mathcal{I}_{|m}$ coincide on the sub-structure rooted at $d'$ and formed by the individuals reached via $\mathtt{member}$ and roles corresponding to attributes, and we are done. Otherwise, in $\mathcal{I}_{|m}$ such sub-structure is expanded into a finite tree. Since by construction the depth of this tree is at least $depth(T)$, and the connections between individuals in $\mathcal{I}$ are preserved in $\mathcal{I}_{|m}$, it follows that $d' \in (\psi(T))^{\mathcal{I}_{|m}}$. $\qquad\square$

The basic reasoning services considered in object-oriented databases are subtyping (check whether a type denotes a subset of another type in every legal instance) and type consistency (check whether a type is consistent in a legal instance). Based on Proposition 5.9, we can show that these forms of reasoning are fully captured by finite concept consistency and finite concept subsumption in ALUNI knowledge bases.

**Theorem 5.10** *Let $\mathcal{S}$ be an object-oriented schema, $T, T'$ two type expressions in $\mathcal{S}$, and $\psi(\mathcal{S})$ the translation of $\mathcal{S}$. Then the following holds:*

*1. $T$ is consistent in $\mathcal{S}$ if and only if $\psi(\mathcal{S}) \not\models_f \psi(T) \preceq \bot$.*

*2. $T$ is a subtype of $T'$ in $\mathcal{S}$ if and only if $\psi(\mathcal{S}) \models_f \psi(T) \preceq \psi(T')$.*

*Proof.* The proof is analogous to the proof of Theorem 4.9, but it makes use of Proposition 5.9 instead of Proposition 4.8. □

Again, the correspondence with ALUNI established by Theorem 5.10 allows us to make use of the reasoning techniques developed for ALUNI to reason on object-oriented schemas. Observe that reasoning in object-oriented models is already PSPACE-hard (Bergamaschi & Nebel, 1994) and thus the known algorithms are exponential. However, by resorting to ALUNI, it becomes possible to take into account for reasoning also various extensions of the object-oriented formalism. Such extensions are useful for conceptual modeling and have already been proposed in the literature (Cattell & Barry, 1997). First of all, the same considerations developed for the ER model with regard to the use of arbitrary boolean constructs on classes can be applied also in the object-oriented setting, which provides disjunction but does not admit any form of negation. Additional features that can be added to object oriented models are inverses of attributes, cardinality constraints on set-valued attributes, and more general forms of restrictions on the values of attributes.

## 6. Related Work

In this section we briefly discuss recent results on the correspondence between class-based formalisms and on techniques for reasoning in ALUNI and in class-based representation formalisms.

### 6.1 Relationships among Class-Based Formalisms

In the past there have been several attempts to establish relationships among class-based formalisms. Bläsius, Hedstück, and Rollinger (1990), Lenzerini, Nardi, and Simi (1991) carry out a comparative analysis of class-based languages and attempt to provide a unified view. The analysis makes it clear that several difficulties arise in identifying a common framework for the formalisms developed in different areas. Some recent papers address this problem. For example, an analysis of the relationships between frame-based languages and types in programming languages has been carried out by Borgida (1992), while Bergamaschi and Sartori (1992), Piza, Schewe, and Schmidt (1992) use frame-based languages to enrich the deductive capabilities of semantic and object-oriented data models.

Artale, Cesarini, and Soda (1996) study reasoning in object-oriented data models by presenting a translation to DLs in the style of the one discussed in Section 5. However, the proposed translation is applicable only in the case where the shema contains no recursive class declarations. This limitation is not present in the work by Bergamaschi and Nebel (1994), where a formalism derived from DLs is used to model complex objects and an algorithm for computing subsumption between classes is provided.

A recent survey on the application of DLs to the problem of data management has been presented by Borgida (1995) . The application to the task of data modeling of reasoning techniques derived from the correspondences presented in Sections 4 and 5 is discussed in more detail by Calvanese, Lenzerini, and Nardi (1998).

Recently, there have also been proposals to integrate the object-oriented and the logic programming paradigms (Kifer & Wu, 1993; Kifer, Lausen, & Wu, 1995). These proposals are however not directly related to the present work, since they aim at providing mechanisms for computing with structured objects, rather than means for reasoning over a conceptual (object-oriented) representation of the domain of interest.

## 6.2 Reasoning in ALUNI and in Class-Based Representation Formalisms

ALUNI is equipped with techniques to reason both with respect to unrestricted and with respect to finite models. We briefly sketch the main ideas underlying reasoning in both contexts. A detailed account of the reasoning techniques has been carried out by Calvanese (1996c).

### 6.2.1 UNRESTRICTED MODEL REASONING

We remind that reasoning on a knowledge base with respect to unrestricted models amounts to check either concept consistency, i.e., determine whether the knowledge base admits a (possibly infinite) model in which a given concept has a nonempty extension, or concept subsumption, i.e., determine whether the extension of one concept is contained in the extension of another concept in every model (including the infinite ones) of the knowledge base.

The method to reason in ALUNI with respect to unrestricted models exploits a well known correspondence between DLs and Propositional Dynamic Logics (PDLs) (Kozen & Tiuryn, 1990), which are a class of logics specifically designed to reason about programs. The correspondence, which has first been pointed out by Schild (1991), relies on a substantial similarity of the interpretative structures of both formalisms, and allows one to exploit the reasoning techniques developed for PDLs to reason in the corresponding DLs. In particular, since $\mathcal{ALUNI}$, the description language of ALUNI, includes the construct for inverse roles, for the correspondence one has to resort to *converse-PDL*, a variant of PDL that includes converse programs (Kozen & Tiuryn, 1990). However, because of the presence of number restrictions in $\mathcal{ALUNI}$ which have no direct correspondence in PDLs, we cannot rely on traditional techniques for reasoning in PDLs. Recently, encoding techniques have been developed, which allow one to eliminate number restrictions from a knowledge base while preserving concept consistency and concept subsumption (De Giacomo & Lenzerini, 1994a). The encoding is applicable to knowledge bases formulated in expressive variants of DLs, and in particular it can be used to reduce unrestricted model reasoning on ALUNI knowledge bases (both concept consistency and concept subsumption) to deciding satisfiability of a formula of converse-PDL. Reasoning in converse-PDL is decidable in EXPTIME (Kozen & Tiuryn, 1990), and since the encoding is polynomial (De Giacomo & Lenzerini, 1994a) we obtain an EXPTIME decision procedure for unrestricted concept consistency and concept subsumption in ALUNI knowledge bases. A simplified form of the encoding, which can be applied to decide unrestricted concept consistency in ALUNI has also been presented by Calvanese et al. (1994).

### 6.2.2 FINITE MODEL REASONING

We remind that reasoning on a knowledge base with respect to finite models amounts to check either finite concept consistency or finite concept subsumption, for which only the *finite* models of the knowledge base must be considered.

For finite model reasoning, the techniques based on a reduction to reasoning in PDLs are not applicable. Indeed, the PDL formula corresponding to an ALUNI knowledge base contains constructs both for converse programs (corresponding to inverse roles) and for functionality of direct and inverse programs, and thus is a formula of a variant of PDL which does not have the finite model property (Vardi, 1985). However, after encoding functionality, one obtains a converse-PDL formula, and since converse-PDL has the finite model property (Fischer & Ladner, 1979), this formula is satisfiable if and only if it is finitely satisfiable. This shows that the encoding of number restrictions (and in particular the encoding of functionality), while preserving unrestricted satisfiability does not preserve finite satisfiability (De Giacomo & Lenzerini, 1994a).

For finite model reasoning in ALUNI one can adopt a different technique, which is based on the idea of separating the reasoning process in two distinct phases (see Calvanese, 1996c, for full details). The first phase deals with all constructs except number restrictions, and builds an "expanded knowledge base" in which these constructs are embedded implicitly in the concepts and roles. In the second phase the assertions involving number restrictions are used to derive from this expanded knowledge base a system of linear inequalities. The system is defined in such a way that its solutions of a certain type (acceptable solutions) are directly related to the finite models of the original knowledge base. In particular, from each acceptable solution one can directly deduce the cardinalities of the extensions of all concepts and roles in a possible finite model. The proposed method allows one to establish for ALUNI EXPTIME decidability for finite concept consistency and for special cases of finite concept subsumption. By resorting to a more complicated encoding one can obtain a 2EXPTIME decision procedure for finite concept subsumption in ALUNI in general (Calvanese, 1996a, 1996c).

Reasoning with respect to finite models has also been investigated in the context of dependency theory in databases. As shown by Casanova, Fagin, and Papadimitriou (1984) for the relational model, when functional and inclusion dependencies interact, the dependency implication problem in the finite case differs from the one in the unrestricted case. While the implication problem for arbitrary functional and inclusion dependencies is undecidable (Chandra & Vardi, 1985; Mitchell, 1983), for functional and unary inclusion dependencies it is solvable in polynomial time, both in the finite and the unrestricted case (Cosmadakis et al., 1990).

Consistency with respect to finite models of schemata expressed in an enriched Entity-Relationship model with cardinality constraints has been shown decidable in polynomial time by Lenzerini and Nobili (1990). Calvanese and Lenzerini (1994b) extend the decidability result to include also ISA relationships, and Calvanese and Lenzerini (1994a) show EXPTIME decidability of reasoning in an expressive object-oriented model. An algorithm for computing a refinement ordering for types (the analogue to a concept hierarchy) in the framework of the $O_2$ object oriented model in discussed by Lecluse and Richard (1989).

Reasoning in the strict sublanguage of ALUNI obtained by omitting inverse roles and number restrictions is already EXPTIME-hard (Calvanese, 1996b). Therefore, the known algorithms for deciding unrestricted concept consistency and subsumption and finite concept consistency are essentially optimal.

## 7. Conclusions

We have presented a unified framework for representing information about class structures and reasoning about them. We have pursued this goal by looking at various class-based formalisms proposed in different fields of computer science, namely frame based systems used in knowledge representation, and semantic and object-oriented data models used in databases, and rephrasing them in the framework of description logics. The resulting description logic, called ALUNI includes a combination of constructs that was not addressed before, although all of the constructs had previously been considered separately.

The major achievement of the paper is the demonstration that class-based formalisms can be given a precise characterization by means of a powerful fragment of first-order logic, which thus can be regarded as the essential core of the class-based representation formalisms belonging to all three families mentioned above. This has several consequences.

First of all, any of the formalisms considered in the paper can be enriched with constructs originating from other formalisms and treated in the general framework. In this sense, the work reported here not only provides a common powerful representation formalism, but may also contribute to significant developments for the languages belonging to all the three families. For example, the usage of inverse roles in concept languages greatly enhances the expressivity of roles, while the combination of ISA, number restrictions, and union enriches the reasoning capabilities available in semantic data models.

Secondly, the comparison of class-based formalisms from the fields of knowledge representation and conceptual data modeling makes it feasible to address the development of reasoning tools to support conceptual modeling (Calvanese et al., 1998). In fact, reasoning capabilities become especially important in complex scenarios such as those arising in heterogenous database applications and Data Warehousing. This line of work was among the motivations for developing systems based on expressive description logics (Horrocks, 1998; Horrocks & Patel-Schneider, 1999), and has lead to further extending the language of description logics to support Information Integration and, more specifically, the conceptual modeling of Data Warehouses (Calvanese, De Giacomo, Lenzerini, Nardi, & Rosati, 1998).